\documentclass[twocolumn]{article}

\usepackage{graphicx}      %
\usepackage{amsmath}       %
\usepackage{amssymb}       %

\usepackage[a4paper,margin=1in]{geometry}
\usepackage{microtype}     %
\usepackage{mathptmx}      %
\usepackage{titlesec}
\usepackage{authblk}

\usepackage[numbers]{natbib} %

\usepackage{booktabs}
\usepackage{multirow}
\usepackage{makecell}
\usepackage{array}
\usepackage{algorithm}
\usepackage{algpseudocode}

\usepackage{xcolor}        %
\usepackage[font=small,labelfont=bf]{caption}

\usepackage[switch]{lineno}      %
\usepackage{threeparttablex}

\usepackage[colorlinks=true, linkcolor=blue, citecolor=blue]{hyperref}
\titleformat{\subsection}[block]
  {\normalfont\normalsize\bfseries}{\thesubsection}{1em}{}

\title{\bfseries Online Continual Learning on Intel Loihi 2 via a Co-designed Spiking Neural Network}

\newcommand{\titleblock}[1]{\twocolumn[\maketitle\begin{abstract}\noindent #1\end{abstract}\vspace{1cm}]}

\begin{document}

\author[1]{Elvin Hajizada$^*$}
\author[2]{Danielle Rager}
\author[3]{Timothy Shea}
\author[2]{Leobardo Campos-Macias}
\author[2]{Andreas Wild}
\author[1,4,5]{Eyke Hüllermeier}
\author[6]{Yulia Sandamirskaya}
\author[2]{Mike Davies}

\affil[1]{Institute of Informatics, University of Munich (LMU), Munich, Germany}  
\affil[2]{Intel Labs, Intel Corporation, Santa Clara, CA, USA}
\affil[3]{Southwest Research Institute, San Antonio, TX, USA}
\affil[4]{Munich Center for Machine Learning (MCML)}
\affil[5]{German Research Centre for Artificial Intelligence (DFKI), Kaiserslautern, Germany}
\affil[6]{Institute of Computational Life Sciences (ICLS), Zurich University of Applied Sciences (ZHAW), Wädenswil, Switzerland}
\affil[*]{Corresponding author: hajizada.elvin@campus.lmu.de}

\renewcommand\Affilfont{\itshape\small}

\date{\today} 

\titleblock{%
Edge AI must adapt to new classes and shifting data within a tight power budget, but online continual learning (learning from each sample before the next arrives) is where conventional processors are least efficient: each small update pays a full dispatch at full power. We present CLP-SNN, a spiking architecture that recasts the Continually Learning Prototypes (CLP) algorithm as event-driven neural dynamics: a self-normalizing local rule replaces CLP's global renormalization, prototype allocation becomes novelty-triggered neurogenesis, and correctness-modulated plasticity becomes reward-gated per-neuron metaplasticity. Deployed on Intel Loihi~2 as the first online continual learner on that chip, rehearsal-free CLP-SNN matches Replay and SLDA on OpenLORIS at 1-shot; at 25-shot, it stays within 0.8 percentage points of Replay and 6.2 points below covariance-based SLDA. Each prediction-and-update step on extracted features completes on-chip in 0.33~ms at 0.05~mJ, $7.3\times$ lower latency and $341\times$ lower energy than Replay on a Jetson Orin Nano.%
}

\section*{Introduction}
\phantomsection\label{sec:intro}

Physical edge AI systems such as service robots, autonomous drones and always-on wearables must operate in environments where lighting, background, or the set of relevant classes can change rapidly and unpredictably~\cite {paleyes2022challenges,jafarzadeh2020openWorldReview,shaheen2022CLrealWorld}. Neural networks are typically trained offline and deployed as static models, under the closed-world assumption that training and deployment distributions remain identical~\cite{reddi2024mlsysbook}; in the open world their accuracy degrades over time~\cite{vela2022AIaging, taori2020imagenetDistShift, xiang2023concept, yuan2022conceptdriftreview}. Periodic retraining and redeployment can address this, but is generally too slow and power-intensive for edge platforms~\cite{khouas2024EdgeTraining}. Large language and vision models offer an alternative to retraining: in-context learning, which adapts by conditioning on demonstrations placed in the model's context without any weight update~\cite{brown2020in_context_learn}. However, these models carry billions of parameters and re-process that context at every query, which makes their deployment on resource-constrained edge devices a significant challenge~\cite{zheng2025edgeLLM}.

Instead, models can adapt continuously to non-stationary data streams on the device~\cite{pellegrini202edge, li2019rilod, ren2021tinyol}. This paradigm, online continual learning (OCL), requires models to predict and learn from each sample once (batch size 1), before the next arrives~\cite{mai2022online,ren2020wandering,hayes2022online,bidaki2025online}. It faces the plasticity-stability dilemma~\cite{kim2023stability}: learning accurately under continual distributional shift while retaining previously acquired knowledge, whose loss is known as catastrophic forgetting~\cite{mccloskey1989catastrophic}. Practical deployments may further require few-shot adaptation from limited examples and the detection of novel concepts~\cite{ren2020wandering,Joseph2021openDetect,jafarzadeh2020openWorldReview,bohus2022wild}. Algorithms and hardware that achieve these capabilities under strict edge constraints have remained elusive.

\begin{figure*}[!h] 
    \centering
    \includegraphics[width=0.995\textwidth]{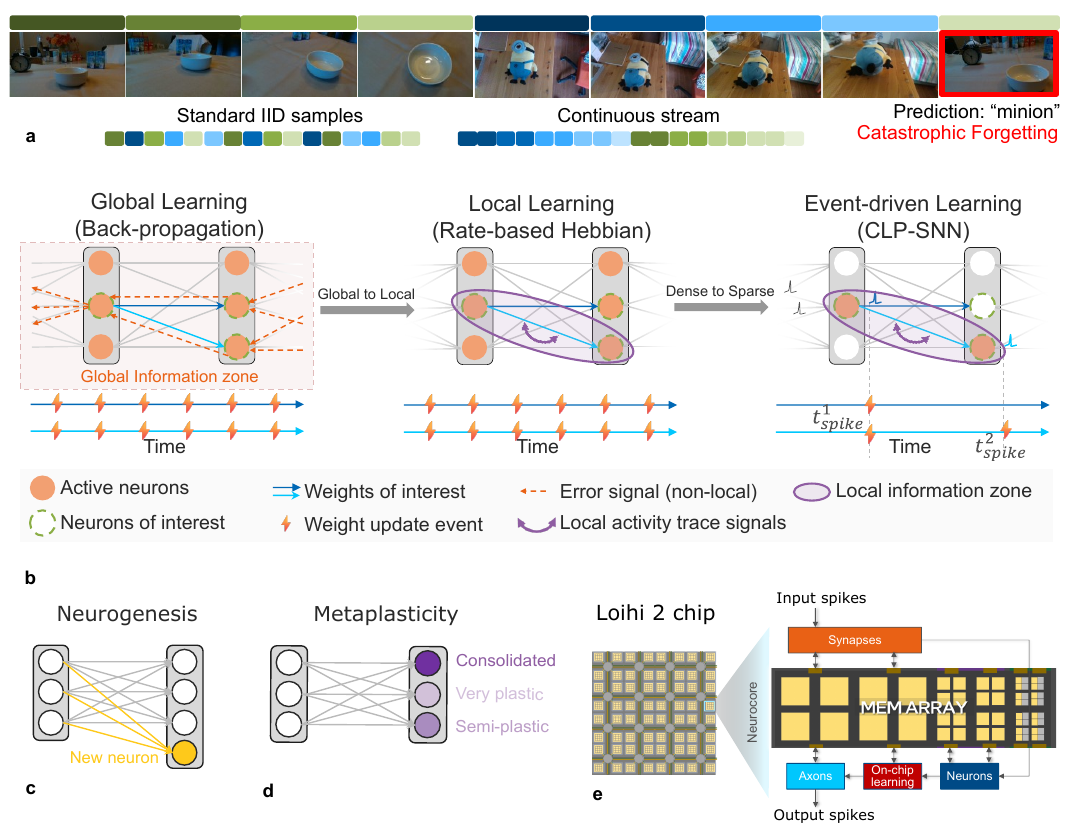}
        \caption{\textbf{From dense global updates to event-driven local learning on the neuromorphic chip Loihi 2, as an efficient solution to online continual learning.} \textbf{a} Catastrophic forgetting happens when learning tasks (e.g., objects) are presented sequentially in non-i.i.d.\ data streams (adapted from~\cite{hajizada2024clp}). Online continual learning (OCL) is such a setting, where inference and learning occur per sample. \textbf{b} (Left) Global learning with backpropagation (BP) relies on non-local error signals (orange dashed arrows), which must be routed back across layers and lock each layer's update until they arrive, and it updates all weights at every step (orange flashes). As tracked here for two weights of interest (blue arrows), the resulting learning is spatially and temporally dense, which is especially inefficient in OCL: at batch size 1, every full update serves a single sample. (Middle) Conventional rate-based local learning (Hebbian) replaces the non-local error signal with local activity trace signals confined to a local information zone. While this eliminates cross-layer routing and backward update locking of BP, the updates are still enacted continuously at each time step, keeping the learning spatially and temporally dense. (Right) In event-driven local learning, spike-timing-dependent plasticity (STDP) rules, including the rule of the proposed architecture, CLP-SNN, restrict weight updates to occur only when and where discrete spikes are emitted (e.g., at $t^1_{\mathrm{spike}}$ for the first weight and $t^1_{\mathrm{spike}}$ and $t^2_{\mathrm{spike}}$ for the second). CLP-SNN's three-factor rule uses only variables local to each synapse, so it can exploit the locality of neuromorphic hardware (\textbf{e}), and a selective modulatory signal triggers it, making updates both spatially and temporally sparse. \textbf{c} CLP-SNN realizes the on-demand prototype allocation of the Continually Learning Prototypes (CLP) algorithm as neurogenesis, recruiting new neurons as new concepts are learned, and \textbf{d} CLP's correctness-modulated learning rate as metaplasticity, which modulates plasticity over time (from very plastic to consolidated) to address catastrophic forgetting. \textbf{e} The Intel Loihi 2 neuromorphic chip, whose neurocores support on-chip, event-driven communication across axons, neurons and synapses, enables local, sparse learning.}
    \label{fig:main_fig}
\end{figure*}

Biological brains face the same problems~\cite{Kudithipudi2022bioCL}: metaplasticity stabilizes memories~\cite{abraham2008metaplasticity, benna2016synConsolidation}, neurogenesis expands capacity on demand~\cite{aimone2009neurogenesis}, synaptic learning rules use only locally available signals~\cite{oja1982simplified, lillicrap2020backpropagation}, and communication is asynchronous and event-driven~\cite{gerstner2014neuronal}. Spiking neural networks (SNNs) offer the same ingredients through sparse, event-driven spikes and Hebbian-like local plasticity~\cite{gerstner2014neuronal, yamazaki2022SNNreview}. Neuromorphic processors such as Intel's Loihi 2 realize them in silicon, integrating memory and compute within each core~\cite{davies2021advancing, mishra2023survey_NC_CL, schuman2022neuromorphicSurvey} and executing local plasticity rules in a dedicated on-chip learning engine~\cite{orchard2021loihi, davies2018loihiLearning}. These features have yielded orders-of-magnitude reductions in latency and energy over CPU and GPU platforms~\cite{orchard2021loihi, davies2021advancing, shrestha2024efficient}.

Yet modern OCL algorithms are not designed for neuromorphic hardware. Most rely on replay buffers to mitigate forgetting. These buffers grow with the number of tasks~\cite{hayes2019protoReplay}, strain the limited on-chip memory of edge devices, and consume energy each time past data are revisited~\cite{verwimp2021rehearsalLimits}. Other methods rely on dense gradient updates or on a full covariance matrix shared across classes, neither of which can be maintained with local operations~\cite{aggarwal2023chameleon, hayes2022online}. Meeting edge constraints therefore calls for co-designing algorithms and hardware~\cite{hajizada2024clp, lin2022device256kb, kukreja2019edgeLearningWhy}. The Continually Learning Prototypes (CLP) algorithm~\cite{hajizada2024clp} is a step in this direction: it is rehearsal-free and updates a set of class prototypes through local Hebbian-like rules. However, each prototype update still concludes with an explicit L2 renormalization of its weight vector, a global gather--norm--broadcast operation that violates on-chip locality and blocks direct deployment on neuromorphic hardware such as Loihi 2.

On the hardware side, online continual learning on neuromorphic chips remains underexplored~\cite{wang2024comprehensiveCL, mishra2023survey_NC_CL, minhas2024review_NC_CL}. Most neuromorphic CL methods are evaluated in software simulation only. The few hardware deployments we know of all ran on the first-generation Loihi: Imam and Cleland~\cite{imam2020epl} employed neurogenesis for continual odor learning, Hajizada et al.~\cite{hajizada2022interactive} designed neural state machines for object learning, and Stewart et al.~\cite{stewart2020online} proposed SOEL for few-shot gesture recognition. None of these has been benchmarked against OCL algorithms running on conventional edge hardware~\cite{hayes2022online}, and OCL studies rarely report throughput or training latency~\cite{mai2022online}.

Here, we introduce CLP-SNN, a spiking neural network architecture that recasts the CLP algorithm as event-driven neural dynamics without the global renormalization identified above, and deploy it on Intel's Loihi 2 neuromorphic processor (Fig.~\ref{fig:main_fig}). Concretely, this work contributes:
(1)~the CLP-SNN architecture, which expresses CLP's control flow as the dynamics of four interacting neuron populations that use only local variables and generic neuromorphic primitives, realizes CLP's prototype allocation as novelty-triggered neurogenesis and its correctness-modulated learning rate as per-neuron metaplasticity, and is released as floating-point and INT8 simulators~\cite{clpsnn_code};
(2)~a self-normalizing three-factor local learning rule that replaces CLP's explicit L2 renormalization by approximating normalized weight updates from only per-synapse variables, so that the weight update maps onto Loihi~2's on-chip learning engine;
(3)~the deployment of CLP-SNN on Intel's Loihi~2, to our knowledge the first online continual learner on that chip, which runs inference on the chip's programmable neuron models, dendritic compartments and graded spikes and the weight update in its learning engine, both on-chip; and
(4)~a cross-platform benchmarking protocol for online continual learners that charges every method for one test-then-train step (prediction plus update) per sample and measures all platforms with the same harness, applied to the OpenLORIS robotic object-recognition dataset~\cite{she2020openloris} with an NVIDIA Jetson Orin Nano as the conventional reference.
On OpenLORIS, rehearsal-free CLP-SNN matches Replay and SLDA at 1-shot and stays close to Replay at 25-shot, where SLDA's covariance model remains more accurate. CLP-SNN completes each prediction-and-update step on the Loihi~2 chip in 0.33~ms at 0.05~mJ. Comparing the same algorithm on the Orin and on Loihi~2 shows that the substrate's contribution is mainly energy.

\begin{figure*}[!t] 
    \centering
    \includegraphics[width=0.995\textwidth]{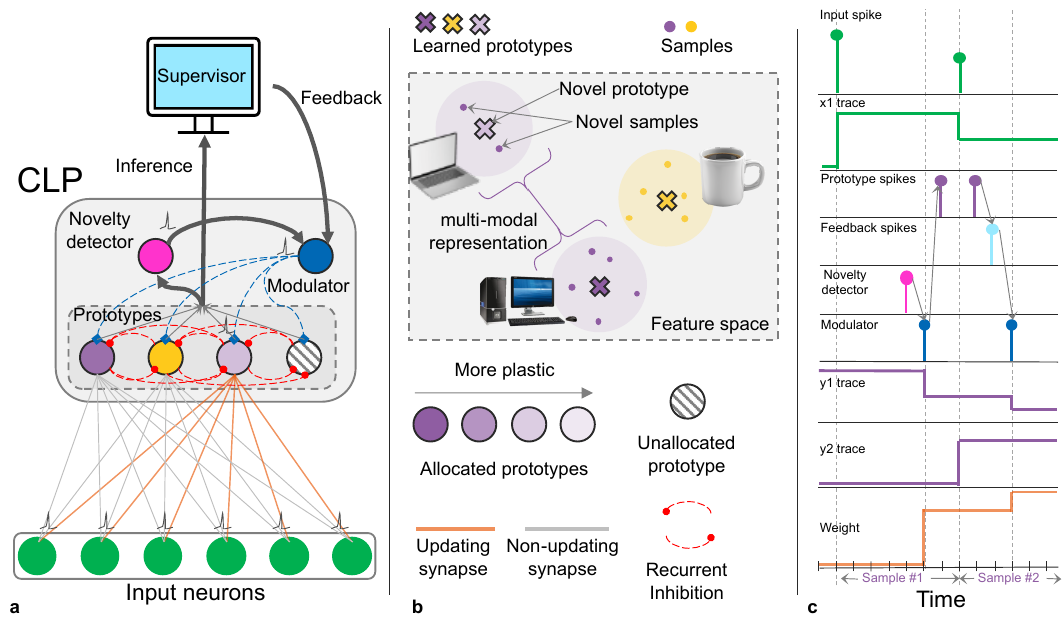}
        \caption{\textbf{The proposed spiking neural network (SNN) architecture for the Continually Learning Prototypes (CLP) algorithm}.~\textbf{a} The SNN consists of multiple neural populations, where the prototype neurons are central. Each prototype neuron has its own plasticity level, which controls the scale of the update applied to its input synapses. The prototypes compete by time-to-first-spike coding, in which the most similar prototype fires first, and lateral inhibition then silences the others (winner-take-all). For each sample, only one (winner) prototype neuron updates its weights. If no neuron spikes, a novelty-detector neuron flags the input as novel and triggers a modulator neuron, which in turn triggers learning in one of the unallocated prototype neurons. If a prototype neuron wins, its label is the prediction, and the supervisor sends a feedback signal based on the correctness of this inference. The modulator neuron relays this signal to the prototype population to update the winning prototype neuron. \textbf{b} A pictographic view of the feature space and representation of the prototypes in this space. Each class may have a variable number of prototype neurons. \textbf{c} Temporal dynamics of inference, novelty detection, feedback and learning at SNN time-step resolution. The first sample of ``laptop'' triggers the allocation of a new (transparent) purple prototype neuron, while the second sample updates this neuron's synapses further. The dynamics are depicted for one of this neuron's input synapses over the processing of two samples.}
    \label{fig:figure_2}
\end{figure*}

\section*{Results}

\subsection*{CLP-SNN expresses prototype-based continual learning as spiking dynamics}

The CLP algorithm~\cite{hajizada2024clp} represents each class by a variable number of \emph{prototypes} in feature space (Fig.~\ref{fig:figure_2}b) and adjusts them online from streaming input. Two mechanisms address catastrophic forgetting without rehearsal or a memory buffer: metaplasticity, which adjusts each prototype's plasticity (learning rate) according to the correctness of inference (Fig.~\ref{fig:main_fig}d), and neurogenesis, which allocates a new prototype whenever an input matches no stored prototype closely enough (novelty detection).

We designed CLP-SNN, a spiking neural network architecture that implements CLP for on-chip continual learning on Loihi~2 (Fig.~\ref{fig:figure_2}). Four interacting neuron populations (input neurons, prototype neurons, a novelty detector and a modulator) implement inference, prototype competition, novelty-triggered neurogenesis and three-factor local weight updates~\cite{fremaux2016threefactor} through their spike-based interactions. All inference and learning computations execute on-chip; per sample, the host compares the predicted label with the ground truth and returns the result to the chip.

Each prototype neuron stores its prototype vector in its input weights (Fig.~\ref{fig:figure_2}a-b). With L2-normalized inputs and weights, as in CLP, the activation a prototype neuron accumulates is the cosine similarity between the input and its prototype. This activation is converted into a spike that occurs earlier the higher the activation, so the earliest spike identifies the winning prototype, and lateral inhibition among prototype neurons (winner-take-all, WTA) keeps a single winner. The winner's spike goes to a \emph{supervisor} as the prediction (Fig.~\ref{fig:figure_2}a) and to a \emph{novelty-detector neuron}, which spikes if no prototype has fired within a set delay after input injection (Fig.~\ref{fig:figure_2}c). The novelty spike activates the \emph{modulator neuron}, whose feedback spike to the prototype population forces a prototype neuron that has not yet learned any pattern to spike and imprint the input as a new prototype (Fig.~\ref{fig:figure_2}c). The modulator neuron also relays the supervisor's positive or negative feedback, obtained by comparing the output with the label, to the winner prototype neuron as a third-factor spike.

For each sample, only the winner prototype neuron learns: the third-factor spike updates its input weights and its individual learning rate, and this per-neuron plasticity implements metaplasticity. Each sample is processed over multiple SNN time steps during which prototype, modulatory and feedback spikes are produced, before the algorithmic time step advances to the next sample (full specification in Methods, \hyperref[sec:methods_architecture]{Architecture}). The Loihi~2 deployment instantiates this architecture with one simplification: each prototype is imprinted once at allocation and not updated afterwards, and a misclassified input allocates a new prototype (Methods, \hyperref[sec:methods_hw_simplification]{``Loihi 2 hardware simplification''}).

\subsection*{CLP-SNN's local learning rule is self-normalizing}

The original CLP algorithm~\cite{hajizada2024clp} keeps each prototype at unit norm by renormalizing its weight vector after every update:
\begin{equation} \label{eq:original_clp_rule}
\mathbf{w}^{(k)}_{n+1} \leftarrow \frac{\mathbf{w}^{(k)}_n + \alpha^{(k)}_n r^{(k)}_n \mathbf{x}_n}{\bigl\|\mathbf{w}^{(k)}_n + \alpha^{(k)}_n r^{(k)}_n \mathbf{x}_n\bigr\|} \, ,
\end{equation}
where, for the $n$-th sample, $\mathbf{x}_n \in \mathbb{R}^d$ is the input vector, $\mathbf{w}^{(k)}_n \in \mathbb{R}^d$ the weight vector of prototype $k$, $\alpha_n^{(k)} \in [0, 1] \subset \mathbb{R}$ its learning rate (i.e., plasticity) and $r^{(k)}_n \in \{-1, 0, +1\}$ the feedback on its prediction: $+1$ if correct, $-1$ if incorrect and $0$ for all non-winner prototypes, so that only the winner learns.

The denominator of Eq.~(\ref{eq:original_clp_rule}) is a global operation: no weight can be rescaled before the whole updated vector is known. Biological neurons lack the machinery for it, and on neuromorphic chips it violates the locality that underlies their energy and latency advantages. On Loihi~2, learning runs on the neurocores, whose learning engines update each synapse from local variables only. Renormalization would instead fall to the chip's few small embedded processors, which serve infrequent management tasks; reading, rescaling and rewriting a whole weight vector at every update would interrupt neurocore processing and add latency.

\begin{figure*}[!h]
    \centering
    \includegraphics[width=0.995\textwidth]{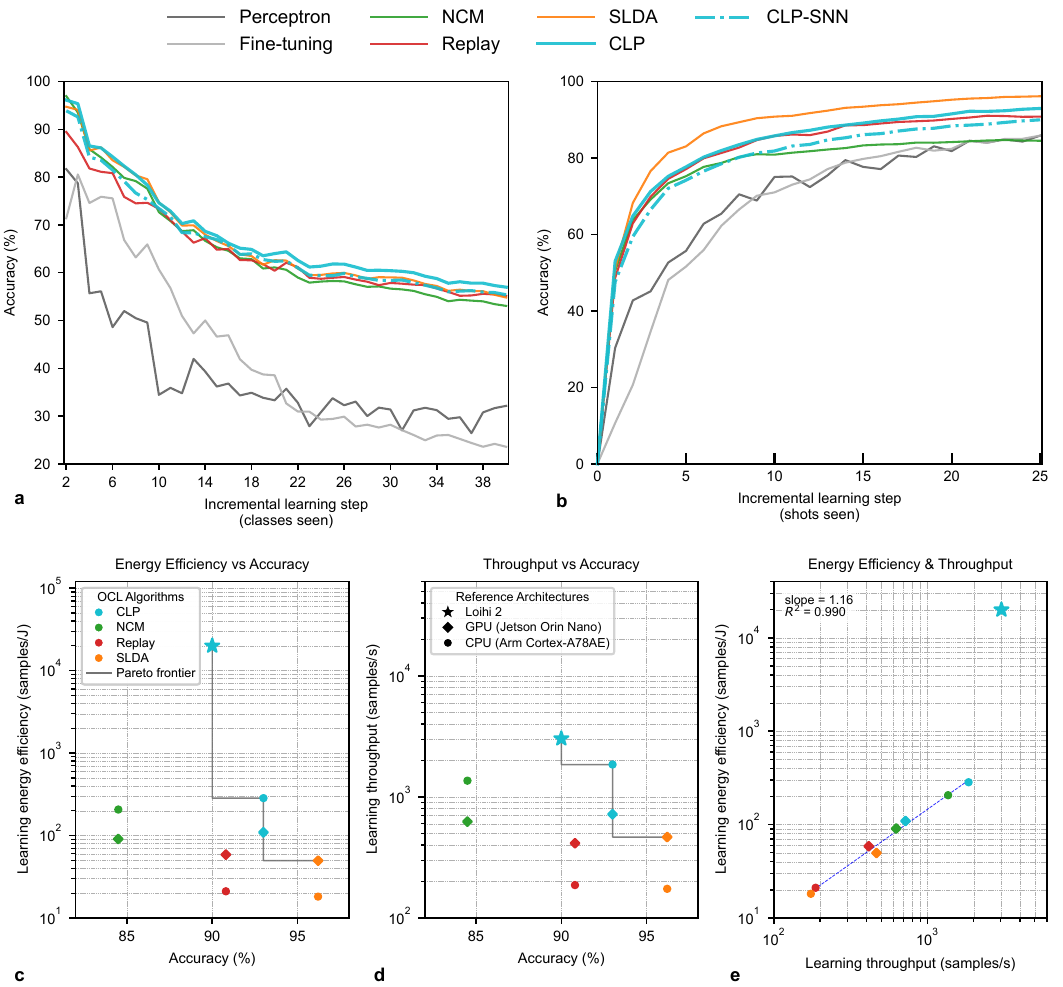}
    \caption{\textbf{Few-shot online continual learning (OCL) experiments on the OpenLORIS dataset.} Acronyms: CLP, Continually Learning Prototypes; CLP-SNN, the proposed CLP spiking neural network; NCM, Nearest Class Mean; SLDA, Streaming Linear Discriminant Analysis. In \textbf{a} and \textbf{b}, lines show the mean over three class orders; s.d.\ of final accuracy is given in Table~\ref{table:1}. \textbf{a} Average accuracy on the classes seen so far during the 1-shot experiment. While the non-continual baselines (grays) fail, OCL algorithms retain accuracy, and CLP outperforms the baselines. The gradual decay of accuracy is a result of the increase in task difficulty (more classes to differentiate among). \textbf{b} 25-shot OCL experiment. Each incremental learning step corresponds to one shot. CLP outperforms all methods except SLDA, which, thanks to its use of the covariance matrix, can leverage the data better, albeit with higher latency and energy costs (see Table~\ref{table:1}). In both experiments, CLP-SNN slightly underperforms CLP (\hyperref[supp:gap]{Supplementary Note~1}). \textbf{c} Learning energy efficiency (samples per joule) versus accuracy at the end of the 25-shot experiment, for each algorithm on each platform (Table~\ref{table:1}); the staircase marks the Pareto frontier. \textbf{d} Learning throughput (samples per second) versus accuracy for the same configurations. \textbf{e} Log--log relationship between learning throughput and energy efficiency. The dashed blue line is a least-squares fit through the conventional hardware configurations (slope = 1.16, $R^2$ = 0.99). The near-unit slope reflects a shared power envelope rather than a constant cost per operation: the operation counts behind these points span two orders of magnitude (Supplementary Table~\ref{tab:op_counts}), but every conventional configuration draws similar power during the step, so energy per sample tracks latency. CLP-SNN on Loihi 2 sits $38\times$ above the line because its per-sample cycle runs at far lower power, not because it performs fewer operations; at the benchmark working point it issues more multiply--accumulate-equivalents per step than CLP on the Orin.}
    \label{fig:exp_results}
\end{figure*}

\begin{figure*}[!h]
    \centering
    \includegraphics[width=0.995\textwidth]{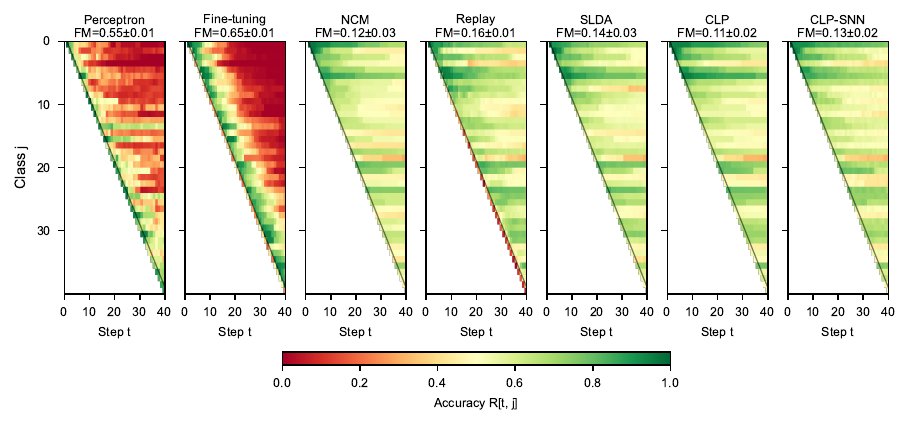}
    \caption{\textbf{Per-class accuracy retention across all 40 classes and 40 learning steps.} Acronyms: OCL, online continual learning; CLP, Continually Learning Prototypes; CLP-SNN, the proposed CLP spiking neural network; FM, forgetting metric. Each heatmap shows $R[t,j]$, the test accuracy on class $j$ after training through step $t$, for one method (seed-averaged), with the black diagonal marking each class's introduction step. Green indicates high accuracy and red low accuracy. Non-continual baselines (Perceptron FM=0.55, Fine-tuning FM=0.65) lose entire rows of early classes to near-zero accuracy within a few steps. OCL methods retain green cells across most of the triangle (CLP FM=0.11, CLP-SNN FM=0.13). The red diagonals of Replay and Fine-tuning mark classes not yet learned rather than forgotten: both learn each class only partly at its introduction and reach their peak accuracy on it in later steps.}
    \label{fig:forgetting_heatmap}
\end{figure*}

A closely related problem was addressed in computational neuroscience by Oja's rule~\cite{oja1982simplified}, which augments the basic Hebbian update~\cite{hebb2005organization} with a multiplicative decay term proportional to the post-synaptic activity. This decay implicitly bounds weight growth and keeps the weight vector approximately normalized without any global operation, using only quantities available at the individual synapse.

We applied the same principle to the CLP learning rule. A Taylor expansion of the renormalized update under a small learning rate (\hyperref[sec:methods_derivation]{derivation in Methods}) yields a local, multiplicative decay term in place of explicit L2 renormalization and produces the self-normalizing CLP-SNN learning rule:

\begin{equation} \label{eq:learning_rule_basic}
\mathbf{w}^{(k)}_{n+1} \leftarrow \mathbf{w}^{(k)}_n + \alpha_n^{(k)} \bigl(r^{(k)}_n \mathbf{x}_n - \bigl|r^{(k)}_n\bigr|\, \mathbf{w}^{(k)}_n y^{(k)}_n\bigr) \, ,
\end{equation}
\begin{equation} \label{eq:post_syn_act}
y^{(k)}_n = \langle\mathbf{w}^{(k)}_n , \mathbf{x}_n\rangle \, ,
\end{equation}
where $y^{(k)}_n \in [-1, 1] \subset \mathbb{R}$ is the post-synaptic activation of prototype $k$, and the feedback $r^{(k)}_n$ reaches the prototype neuron as a third-factor signal. The term $-\bigl|r^{(k)}_n\bigr|\,\mathbf{w}^{(k)}_n y^{(k)}_n$ is the local multiplicative decay that replaces explicit renormalization. The sign of the reward multiplies only the Hebbian term $r^{(k)}_n\mathbf{x}_n$, while its magnitude $|r^{(k)}_n|$ gates the decay to the learning prototype. The Taylor expansion would also apply the sign to the decay, but that would make the unit norm unstable under incorrect predictions: a slightly too long prototype grows with every error (\hyperref[sec:methods_stability]{Methods}). Supplementary Figs.~\ref{fig:learning_rule_input_d_sensitivity} and~\ref{fig:learning_rule_sigma_analysis} characterize the convergence and weight-norm stability of CLP and of CLP-SNN in 64-bit floating point and in INT8 across a range of input dimensionalities and intra-class variance levels, confirming that the self-normalizing rule retains usable prototype alignment across all tested conditions.

\subsection*{CLP-SNN matches replay-based accuracy in few-shot streaming}

Throughout this work we operate in the strict OCL setting at batch size~1, the standard OCL benchmark protocol~\cite{mai2022online,hayes2022online} and the most demanding real-time constraint. We used the OpenLORIS dataset~\cite{she2020openloris}: a single pass over the unshuffled frames of 40 classes, pre-processed by a fixed pre-trained feature extractor on a GPU (Fig.~\ref{fig:main_fig}a).

We compared CLP and CLP-SNN (on Loihi~2) with three OCL baselines~\cite{hayes2022online, hajizada2024clp}: nearest class mean (NCM), with one running-average prototype per class; Replay, which rehearses stored samples; and streaming linear discriminant analysis (SLDA), which adds a shared covariance matrix. Fine-tuning and Perceptron serve as non-continual baselines (\hyperref[sec:methods_baselines]{Methods}). In the 1-shot setting, each method learned a sequence of 40 online learning tasks, each containing a single short video clip per object category, similar to class-incremental learning. After each task, we tested on all classes seen so far, for three random class orders (Fig.~\ref{fig:exp_results}a). CLP attains the highest accuracy, and CLP-SNN matches Replay and SLDA (Table~\ref{table:1}). CLP-SNN on Loihi~2 is slightly less accurate than CLP. The gap comes mainly from the deployed network's lower similarity threshold, which reduces the number of prototypes (and with them the latency and energy per sample), and partly from resolving the winner in discrete spike-time steps (\hyperref[supp:gap]{Supplementary Note~1}).

The heatmaps of per-class accuracy retention in Fig.~\ref{fig:forgetting_heatmap} quantify forgetting using the Forgetting Metric (FM)~\cite{chaudhry2018riemannianWalk}, defined per class as the drop from a class's peak accuracy to its final accuracy and averaged across classes (lower is better; definition, per-class values and temporal evolution in Supplementary Fig.~\ref{fig:detailed_forgetting}). Non-continual baselines (Perceptron FM=0.55, Fine-tuning FM=0.65) lose entire rows of early classes within a few steps, while CLP (FM=0.11) and CLP-SNN (FM=0.13) retain accuracy across all 40 classes and 40 learning steps.

\begin{table*}[th]
\centering
\begin{threeparttable}
\small %
\caption{Accuracy and per-sample cost of one online continual learning
step. The step comprises producing the prediction for the incoming sample
and then updating the model on its label, the test-then-train order in
which an online learner meets every sample. It runs at batch size~1, with each
method in an efficient implementation of the same step; feature
extraction and host--device I/O are excluded on all platforms. For CLP and
Replay the update pass already computes the prediction (winner selection and
the training forward pass, respectively), so the step is the update alone;
for NCM and SLDA the update does not score the sample, so the step is the
sum of the two primitives. On Loihi~2 the step is a single on-chip pass. The
per-primitive breakdown, the half-precision variants and an
explicit-inversion SLDA are reported in
Supplementary Table~\ref{tab:cost_primitives}. Accuracies are final values,
mean\,$\pm$\,s.d.\ over three class orders. Best value per column in bold,
second best underlined.}
\label{table:1}
\begin{tabular}{lccccccc}
\toprule
\textbf{Method} & \textbf{System} & \textbf{Quantization} & \makecell{\textbf{Acc (\%) }\\\textbf{(1-shot)}}   & \makecell{\textbf{Acc (\%) }\\\textbf{(25-shot)}} & \makecell{\textbf{Latency}\\\textbf{(ms)}}  & \makecell{\textbf{Dynamic}\\\textbf{energy (mJ)}} & \makecell{\textbf{Total}\\\textbf{energy (mJ)}} \\
\midrule
\multirow{1}{*}{CLP-SNN} & Loihi 2\tnote1 & INT8 & \underline{55.4}{\scriptsize$\pm$2.5} & 90.0{\scriptsize$\pm$0.4} & \textbf{0.33} & \textbf{0.01} & \textbf{0.05} \\  %
\midrule
\multirow{2}{*}{CLP} & GPU\tnote2 & FP32 & \textbf{57.0}{\scriptsize$\pm$2.4} & \underline{93.0}{\scriptsize$\pm$0.2} & 1.39 & 1.15 & 9.15 \\  %
 & CPU\tnote3 & FP32 & \textbf{57.0}{\scriptsize$\pm$2.4} & \underline{93.0}{\scriptsize$\pm$0.2} & \underline{0.54} & \underline{0.41} & \underline{3.52} \\  %
\midrule
\multirow{2}{*}{NCM} & GPU & FP32 & 53.1{\scriptsize$\pm$2.9} & 84.5{\scriptsize$\pm$0.8} & 1.60 & 1.79 & 11.02 \\  %
 & CPU & FP32 & 53.1{\scriptsize$\pm$2.9} & 84.5{\scriptsize$\pm$0.8} & 0.73 & 0.61 & 4.84 \\  %
\midrule
\multirow{2}{*}{Replay} & GPU & FP32 & 55.1{\scriptsize$\pm$2.8} & 90.8{\scriptsize$\pm$0.6} & 2.41 & 3.11 & 17.04 \\  %
 & CPU & FP32 & 55.1{\scriptsize$\pm$2.8} & 90.8{\scriptsize$\pm$0.6} & 5.35 & 16.34 & 47.47 \\  %
\midrule
\multirow{2}{*}{SLDA} & GPU & FP32 & 54.8{\scriptsize$\pm$2.5} & \textbf{96.2}{\scriptsize$\pm$0.1} & 2.15 & 7.34 & 20.17 \\  %
 & CPU & FP32 & 54.8{\scriptsize$\pm$2.5} & \textbf{96.2}{\scriptsize$\pm$0.1} & 5.74 & 20.66 & 55.04 \\  %
\bottomrule
\end{tabular}
\begin{tablenotes}
\footnotesize  %
\item[1]Loihi 2 Oheo Gulch system with N3C2-revision chips running on Lava 0.11.0.dev0, Lava-Loihi 0.6.0.dev0 and NxCore 2.5.8.
\item[2]NVIDIA Jetson Orin Nano 8~GB: JetPack SDK 6.2.1 (15~W TDP), PyTorch 2.4.0.
\item[3]NVIDIA Jetson Orin Nano's 6-core Arm Cortex-A78AE CPU.
\end{tablenotes}
\end{threeparttable}
\end{table*}

\subsection*{CLP-SNN improves the accuracy--latency--energy trade-off}
To probe longer-term stability, we ran a 25-shot protocol: 25 successive learning shots, each providing one video per object, 40 videos per shot. Frames are again unshuffled and seen in a single pass. This allows us to test the models under domain shifts, similar to a domain-incremental setting, as no new class is introduced after the first shot. 

Over the 25 shots the methods separate: CLP surpasses both NCM and Replay, while CLP-SNN ends 3.0 percentage points below CLP and 0.8 points below Replay (Fig.~\ref{fig:exp_results}b, Table~\ref{table:1}). NCM is the simplest form of CLP, with a single prototype per class, so CLP's adaptive multi-prototype model is what leverages the extra data. CLP stores no past samples, so its lead over Replay shows that rehearsal-free learning need not trail rehearsal-based learning on this benchmark. SLDA, in contrast, outperforms CLP by 3.2 percentage points in the 25-shot protocol (96.2\% versus 93.0\%), attributable to its additional modeling capacity. It classifies with the inverse of its shared covariance matrix, the precision matrix, which it updates per sample by an exact rank-one update at a cost of order $d^{2}$ (Methods, \hyperref[sec:methods_rankone_slda]{``Exact rank-one streaming LDA baseline''}).

On Loihi~2, CLP-SNN completes one test-then-train step in 0.33~ms at 0.05~mJ (Table~\ref{table:1}); every baseline on the Jetson Orin Nano is charged for the same step, in an efficient implementation. Against SLDA on the Orin GPU, the most accurate baseline, this is $6.5\times$ lower latency and $403\times$ lower energy, for 6.2 percentage points of 25-shot accuracy; against Replay on the GPU, the closest baseline in accuracy, $7.3\times$ and $341\times$ for 0.8 points.

Those ratios compare different algorithms as well as different substrates. To isolate what the co-design itself buys, we compare CLP-SNN on Loihi~2 with CLP on the Orin. Both classify each sample with one similarity sweep over stored prototypes and write at most one prototype. CLP-SNN replaces explicit renormalization with the local self-normalizing update, uses INT8 arithmetic and selects the winner by spike timing instead of an argmax. On this comparison Loihi~2 delivers $4.2\times$ lower latency and $183\times$ lower energy per learning step than the Orin GPU, and $1.6\times$ lower latency and $70\times$ lower energy than the Orin CPU. The advantage is mainly in energy rather than in speed.

Counting operations shows that these gains do not come from doing less arithmetic (Supplementary Table~\ref{tab:op_counts}). At the benchmark working point the conventional learners span two orders of magnitude in arithmetic per step, from $5.4\times10^{4}$ multiply--accumulate operations for NCM to $5.2\times10^{6}$ for Replay, yet their GPU step latencies differ by at most $1.7\times$ (Table~\ref{table:1}). At batch size~1 the GPU step is set by kernel dispatch rather than by the arithmetic it performs; on the CPU, by contrast, the same range spreads latency by about $11\times$. Lowering the arithmetic precision confirms this directly: in half precision (FP16) no step became faster and no step cost changed by more than 6\% (Supplementary Table~\ref{tab:cost_primitives}). Nor can these learning steps run in INT8 on the Orin, whose INT8 path accelerates only dense matrix multiplication at large batch sizes (\hyperref[sec:methods_benchmarking]{Methods}). Loihi~2 in turn issues about $5.6\times10^{5}$ multiply--accumulate-equivalents per step, more than six times the count of CLP on the Orin, because every input spike fans out to all 300 prototype slots~\cite{timcheck2026runtime} and the plasticity engine sweeps all $1280\times300$ prototype synapses. Despite this larger operation count, Loihi~2 completes the step faster and with far less energy than CLP on the Orin, because each operation runs next to the memory that holds its weight, at low power and with no dispatch to pay.

On conventional hardware alone, the algorithms trade accuracy against energy efficiency (Fig.~\ref{fig:exp_results}c) and against learning throughput (Fig.~\ref{fig:exp_results}d). With Loihi~2 added, the Pareto frontier over all configurations consists of three points: CLP-SNN on Loihi~2, CLP on the Orin CPU and SLDA on the Orin GPU. Across the conventional configurations, energy efficiency and throughput are tightly coupled ($R^{2}=0.99$ on log--log axes, Fig.~\ref{fig:exp_results}e), and CLP-SNN on Loihi~2 sits $38\times$ above that line on total energy because its power envelope is far lower.

\subsection*{Not all sparsity types contribute equally}
On Loihi~2, CLP-SNN processes each sample over a set number of SNN time steps, 20 in our deployment. Each step runs a spiking phase that delivers spikes and updates neuron state; every $t_{\mathrm{epoch}}$-th step also runs a learning phase, in which the plasticity engine sweeps the prototype synapses. The benchmarked configuration exploits three kinds of sparsity: input sparsity, with 50\%-sparse input vectors, close to the mean sparsity of the OpenLORIS features; temporal sparsity of learning, with one learning phase per sample ($t_{\mathrm{epoch}}=20$); and spatial sparsity of learning, with a post-synaptic gate that lets only the winner update.

Ablating each in turn separates their contributions. Input sparsity shortens the spiking phase by about 28\% relative to dense vectors. Temporal sparsity of learning is the dominant lever, shortening the learning phase by nearly 20$\times$ and the sample as a whole by 5.6$\times$, after which the spiking phase rather than learning accounts for most of the sample. Spatial sparsity of learning saves no time: updating only the winner takes as long as updating every prototype. Whenever the learning phase runs, the engine reads every plastic (learning-enabled) synapse, because on Loihi~2 only pre-synaptic events condition which synapses are read. A post-synaptic term gates what the rule computes, zeroing the weight change that would have followed (Supplementary Table~\ref{tab:op_counts}). Sparsity therefore saves time on Loihi~2 only where it removes events, such as input spikes or learning phases, not where it zeroes updates after the synapses have been read.

\begin{table}[h]
\centering
\small
\caption{Learning-step share of the per-sample pipeline cost
(feature extractor\,+\,learning step) on the Jetson Orin Nano GPU,
for the FP32 and INT8 EfficientNet-B0 extractors (TensorRT, batch
size~1), whose own cost per sample is given in the first row.
Step costs are the GPU rows of Table~\ref{table:1}. Dynamic rather
than total energy is shown because static power belongs to the
device and is shared with any concurrent workload.}
\label{tab:ocl_portion}
\begin{tabular}{@{}lrrrr@{}}
\toprule
 & \multicolumn{2}{c}{\textbf{Latency}} &
   \multicolumn{2}{c}{\textbf{Dyn.\ energy}} \\
\cmidrule(lr){2-3}\cmidrule(lr){4-5}
 &
  \textbf{FP32} & \textbf{INT8} &
  \textbf{FP32} & \textbf{INT8} \\
\midrule
Extractor  & 4.8~ms & 2.4~ms & 24.0~mJ & 6.7~mJ \\  %
\midrule
\multicolumn{5}{@{}l}{\emph{Learning-step share}} \\
NCM       & 25\% & 40\% &  7\% & 21\% \\  %
CLP        & 22\% & 37\% &  5\% & 15\% \\  %
Replay     & 33\% & 50\% & 11\% & 32\% \\  %
SLDA       & 31\% & 47\% & 23\% & 52\% \\  %
\bottomrule
\end{tabular}
\end{table}

\subsection*{Learning-step cost within the full pipeline}
Table~\ref{table:1} excludes feature extraction, so its gains are learning-step gains, whose weight in the whole per-sample pipeline depends on the feature extractor. Our experiments use a frozen EfficientNet-B0, a common edge extractor~\cite{hayes2022online}, for all methods; the shares below hold for this extractor on the Orin GPU (Table~\ref{tab:ocl_portion}). With the FP32 extractor, the extractor dominates: the learning step of the conventional methods accounts for 22--33\% of pipeline latency and 5--23\% of pipeline dynamic energy. Quantizing the extractor to INT8 cuts its dynamic energy by 3.6$\times$ and raises these shares to 37--50\% and 15--52\% (Table~\ref{tab:ocl_portion}): with an efficient extractor, the learning step accounts for up to half of the pipeline cost, enough to warrant optimizing it in its own right.

\section*{Discussion}
This work makes four contributions. The CLP-SNN architecture expresses CLP's control flow, including its on-demand prototype allocation and correctness-modulated plasticity, as the dynamics of neuron populations that use only local variables and generic neuromorphic primitives. Its self-normalizing rule removes CLP's global renormalization, so the weight update runs in Loihi~2's learning engine. The Loihi~2 implementation learns in real time at 0.05~mJ per step and mitigates catastrophic forgetting on non-i.i.d.\ data streams, and our benchmarking protocol, which charges every method for the same test-then-train step, makes these measurements reference points for neuromorphic continual learning.

Comparing the implementation on Loihi 2 with CLP on the Orin, which runs the same algorithm, shows that its advantage lies mainly in energy: Loihi~2 lowers the energy per step by 70$\times$ and 183$\times$ against the Orin CPU and GPU, but the latency by only 1.6$\times$ and 4.2$\times$. Every conventional configuration draws similar power during the learning step, whereas Loihi~2 runs the step at far lower power. The chip runs at such low power because each core updates 8-bit weights in logic next to the memory that holds them and, during inference, reads synaptic weights only for inputs that spike. CLP-SNN is designed to exploit both. Its learning rule and neuron dynamics need only local variables, so that inference and the weight update run on-chip. Its input layer emits a spike only for non-zero features. It also runs its learning phase only once per sample rather than at every SNN time step, and of the forms of sparsity we tested, this one shortens the step on Loihi~2 the most.

We note four limitations of this work. First, the Loihi~2 implementation imprints each prototype once at allocation without adapting it afterwards, and therefore requires about twice as many prototypes as a fully adaptive CLP-SNN, which currently runs only in simulation (\hyperref[supp:adaptive]{Supplementary Note~2}). This simplification does not lower the reported step cost: a weight update on the chip costs the same whether it imprints a prototype or adapts one.

Second, the feature extractor is fixed and shared by all compared methods, whereas substantial domain shifts may require the features to adapt as well. Here learning adapts only the final layer and already accounts for up to half of the per-sample cost (Table~\ref{tab:ocl_portion}); adapting more layers would raise that share further, making efficient on-device learning more critical still. A third limitation is scale: the all-to-all lateral inhibition that selects the winning prototype needs $O(K^2)$ connections for $K$ prototypes and will exceed per-core fan-in and fan-out axon limits as prototype populations grow to thousands of neurons. A hub-and-spoke topology, which routes all prototype outputs to a single hub neuron that broadcasts the inhibition, would reduce this to two $O(K)$ fan-out stages natively supported by Loihi~2's axon routing. Finally, CLP-SNN relies on Loihi~2's native graded spikes~\cite{orchard2021loihi}, which carry an integer value per event; with binary spikes, each input value would need several time steps.

Our sparsity ablations also inform the design of next-generation neuromorphic architectures. Because the post-synaptic gate applies only after the learning engine has read every plastic synapse, updating only the winning prototype saved no time. This suggests that future on-chip learning engines should condition strictly on all gating events (pre-synaptic, post-synaptic and third-factor) to support spatiotemporally sparse learning.

\section*{Methods}

\subsection*{Derivation of the CLP-SNN rule}
\phantomsection\label{sec:methods_derivation}

We derive the CLP-SNN rule from the renormalized CLP update, Eq.~(\ref{eq:original_clp_rule}). Only the winning prototype learns, so $r\in\{-1,+1\}$ and $r^2=1$; we drop the indices $k$ and $n$, write $y=\mathbf{w}^\top\mathbf{x}$ for the post-synaptic activation and assume a small, positive learning rate, $0<\alpha\ll 1$. CLP adds the Hebbian term $\alpha r\mathbf{x}$ and renormalizes:
\[
\mathbf{w}^{\text{new}} = \frac{\mathbf{w} + \alpha r\mathbf{x}}{\|\mathbf{w} + \alpha r\mathbf{x}\|} \, .
\]
Assuming that the weight and input vectors have unit norm before the update, $\|\mathbf{w}\|=\|\mathbf{x}\|=1$, the squared norm in the denominator is
\begin{align*}
\|\mathbf{w} + \alpha r\mathbf{x}\|^2 &= \|\mathbf{w}\|^2 + 2\alpha r\,\mathbf{w}^\top\mathbf{x} + \alpha^2 r^2\|\mathbf{x}\|^2 \\
&= 1 + 2\alpha r y + \alpha^2 \, .
\end{align*}
We apply the Taylor expansion of $\sqrt{1 + \xi}$ around $\xi=0$:
\[
\sqrt{1 + \xi} = 1 + \frac{\xi}{2} - \frac{\xi^2}{8} + O(\xi^3) \, .
\]
Setting $\xi = 2\alpha r y + \alpha^2$, so that $\xi^2 = O(\alpha^2)$,
\[
\sqrt{1 + 2\alpha r y + \alpha^2} = 1 + \alpha r y + O(\alpha^2) \, .
\]
Dropping $O(\alpha^2)$ for $\alpha \ll 1$, the renormalized weight vector is
\[
\mathbf{w}^{\text{new}} \approx \frac{\mathbf{w} + \alpha r\mathbf{x}}{1 + \alpha r y} \, .
\]
Applying $(1+\epsilon)^{-1} = 1 - \epsilon + O(\epsilon^2)$ with $\epsilon = \alpha r y$,
\[
\mathbf{w}^{\text{new}} \approx (\mathbf{w} + \alpha r\mathbf{x})(1 - \alpha r y) \, .
\]
Expanding and dropping the $O(\alpha^2)$ term $-\alpha^2 r^2 y\,\mathbf{x}$ yields CLP's update to first order in $\alpha$:
\[
\Delta\mathbf{w} = \mathbf{w}^{\text{new}} - \mathbf{w} \approx \alpha r\,(\mathbf{x} - \mathbf{w}y) \, ,
\]
a Hebbian term $\alpha r\mathbf{x}$ plus a multiplicative decay $-\alpha r\,\mathbf{w}y$. Both are local: each weight $w_j$ changes using only its own value, its pre-synaptic input $x_j$, the post-synaptic activation $y$ and the third factor $r$. Because ${\mathbf{w}^\top(\mathbf{x}-\mathbf{w}y) = y - y = 0}$, the update is orthogonal to $\mathbf{w}$, so to first order CLP only rotates the prototype: toward the input for $r=+1$ and away from it for $r=-1$. The decay term only cancels the change in length that the Hebbian term would cause.

\phantomsection\label{sec:methods_stability}\textbf{Stability of the weight norm.} The derived update preserves the unit norm only to first order and only if $\|\mathbf{w}\|=1$ before the update. The neglected $O(\alpha^2)$ terms and INT8 rounding on the chip move the norm away from one at every update. We therefore analyze the stability of the norm: whether the rule pulls a drifted norm back to one over successive updates. We consider local rules of the form
\[
\Delta\mathbf{w} = \alpha\,(r\,\mathbf{x} - \kappa\,\mathbf{w}y) \, ,
\]
where $\kappa$ is the decay coefficient; the derived update is the case $\kappa=r$. The decay term $\kappa\,\mathbf{w}y$ points along $\mathbf{w}$, so the choice of $\kappa$ changes only the prototype's length: at $\|\mathbf{w}\|=1$, every $\kappa$ rotates the prototype as CLP does, to first order. Let $\rho=\|\mathbf{w}\|$, no longer assumed to be one. Since $\mathbf{w}^\top\mathbf{x}=y$ and $\mathbf{w}^\top\mathbf{w}=\rho^2$, the squared norm changes by
\begin{align*}
\|\mathbf{w}+\Delta\mathbf{w}\|^2 - \rho^2 &= 2\,\mathbf{w}^\top\Delta\mathbf{w} + \|\Delta\mathbf{w}\|^2 \\
&= 2\alpha y\,(r - \kappa\rho^2) + O(\alpha^2) \, ,
\end{align*}
where $y\geq\theta>0$ for the winning prototype, and $\theta$ is the similarity threshold (Eq.~\ref{eq:novelty_threshold}). We measure the drift by the deviation of the squared norm from one: $\delta=\rho^2-1$ before the update and $\delta^{\text{new}}=\|\mathbf{w}+\Delta\mathbf{w}\|^2-1$ after it. Substituting $\rho^2=1+\delta$ into the norm change gives
\[
\delta^{\text{new}} = (1-2\alpha\kappa y)\,\delta + 2\alpha y\,(r-\kappa) + O(\alpha^2) \, .
\]
Each update multiplies the previous deviation by the factor $1-2\alpha\kappa y$, which is smaller than one for $\kappa>0$ and larger than one for $\kappa<0$, and adds the offset $2\alpha y\,(r-\kappa)$, which vanishes only for $\kappa=r$.

The derived update, $\kappa=r$, gives $\delta^{\text{new}} = (1-2\alpha r y)\,\delta + O(\alpha^2)$. For a correct prediction ($r=+1$), each update shrinks the deviation, so the unit norm is an attracting state, the self-normalizing property of Oja's rule~\cite{oja1982simplified}. For an incorrect prediction ($r=-1$), the factor becomes $1+2\alpha y$: the unit norm is repelling, and a prototype that is slightly too long grows with every incorrect prediction.

CLP-SNN therefore sets $\kappa=|r|$, Eq.~(\ref{eq:learning_rule_basic}), which is $\kappa=1$ for the learning prototype: the sign of the reward reaches only the Hebbian term, and its magnitude only gates the decay by whether the prototype learns at all. This gives $\delta^{\text{new}} = (1-2\alpha y)\,\delta + 2\alpha y\,(r-1) + O(\alpha^2)$: the factor is $1-2\alpha y$ for both reward signs, so every update shrinks the previous deviation. A correct prediction adds no offset, and an incorrect one shortens the prototype. For $r=-1$, no $\kappa$ avoids both problems: the offset vanishes only for $\kappa=-1$, which makes the factor exceed one.

After an incorrect prediction, CLP-SNN thus rotates the prototype as CLP does, to first order, but also reduces its length $\rho$ by about $2\alpha y$. The shorter prototype responds more weakly until subsequent correct predictions, which attract the norm back to one, restore its length.

The analysis assumes $\alpha\ll 1$. The learning rate $\alpha=1/g$ (Eq.~\ref{eq:alpha_trace}) equals one at allocation, where the update imprints the input onto zero weights and yields a unit norm exactly, and remains moderate during a prototype's first few updates. There the norm overshoots and returns to one as $\alpha$ decays (Supplementary Fig.~\ref{fig:learning_rule_input_d_sensitivity}).

\subsection*{CLP-SNN architecture: neuron populations and learning}
\phantomsection\label{sec:methods_architecture}

CLP-SNN comprises four spiking neuron populations: input, prototype, novelty-detector and modulator. Their state dynamics and inter-population connections together implement inference, winner-take-all (WTA) competition, novelty-triggered prototype allocation and three-factor local weight updates (Fig.~\ref{fig:figure_2}). The novelty-detector and modulator populations produce the third-factor spike $r^{(k)}_n$ of Eq.~(\ref{eq:learning_rule_basic}), which gates plasticity and triggers neurogenesis. We specify each neuron model below, followed by the WTA competition and tie-break rule and the modulatory-signal logic.

We use two notions of time. An \textbf{algorithmic time step} $n$ processes one sample, including inference, novelty detection, feedback and learning. It is made up of \textbf{SNN time steps} $t$, the discrete cycles in which every neuron updates its state once and spikes are exchanged; unless stated otherwise, one algorithmic time step spans 20 SNN time steps. Equations indexed by $n$ describe per-sample updates, and variables indexed by $t$ resolve the spike dynamics within a sample.

\textbf{Input neuron.} Each input neuron is a stateless graded-spike relay for one feature dimension: on receipt of an input value, it emits a graded spike with integer payload. Because Loihi~2 connections from an external injection process carry no learnable weights, this relay layer is necessary. Graded spikes deliver the full 1280-dimensional feature vector to the prototype population in a single SNN time step, rather than across $T$ rate-coded time steps; because every SNN time step on Loihi~2 ends with a barrier synchronization across cores, fewer time steps per sample mean lower latency.

\textbf{Prototype neuron.} Each prototype neuron is a custom spiking neuron that converts its feedforward activation into a spike time. It has three dendritic accumulator compartments (DA): a \emph{feedforward channel} (DA1) that accumulates graded input spikes via learnable synaptic weights; a \emph{modulatory channel} (DA2) that receives third-factor reward and allocation signals from the modulator neuron; and a \emph{lateral-inhibition channel} (DA3) that receives winner-take-all reset signals from other prototype neurons. Each neuron stores a unique identifier $\mathrm{id}^{(k)}$, a membrane potential $v^{(k)}(t)$, a sample-gated bias current $u^{(k)}(t)$, an adaptive learning rate $\alpha^{(k)}(t)$, and a goodness score $g^{(k)}(t)$; the spike threshold $v_{\mathrm{th}}$ and bias increment $\beta$ are fixed parameters shared across all prototype neurons.

At $t=0$, the input neurons fire simultaneously; the feedforward channel (DA1) accumulates their weighted sum over the synaptic weights, making the dot product $y^{(k)} = \sum_j w^{(k)}_j x_j$ available at $t=1$. This value is $y^{(k)}_n$ of Eq.~(\ref{eq:post_syn_act}), the cosine similarity for L2-normalized vectors, and corresponds to the post-synaptic trace of Eq.~(\ref{eq:loihi2_rule}). The bias $\beta$ is refreshed whenever $y^{(k)}(t) \neq 0$ and held constant for subsequent SNN time steps until the inter-sample reset:
\begin{equation}\label{eq:proto_bias_dynamics}
u^{(k)}(t) = \begin{cases}
\beta & \text{if } y^{(k)}(t) \neq 0, \\
0     & \text{if reset}(t), \\
u^{(k)}(t-1) & \text{otherwise.}
\end{cases}
\end{equation}
The membrane potential integrates the feedforward input and the bias:
\begin{equation}\label{eq:proto_lif_dynamics}
v^{(k)}(t) = v^{(k)}(t-1) + y^{(k)}(t) + u^{(k)}(t).
\end{equation}
Because $v^{(k)}(0)=0$ and $y^{(k)}(t)$ is non-zero only at $t=1$, the potential ramps linearly, $v^{(k)}(t) = y^{(k)} + t\beta$, until a reset sets $v^{(k)}$ and $u^{(k)}$ to zero. With $v_{\mathrm{th}}$ and $\beta$ shared across all prototypes, this ramp gives the spike firing time directly:
\begin{equation}\label{eq:ttfs}
t^{(k)}_s = \left\lceil \frac{v_{\mathrm{th}} - y^{(k)}}{\beta} \right\rceil.
\end{equation}
This implements \emph{time-to-first-spike (TTFS)} coding: the prototype with the highest cosine similarity $y^{(k)}$ fires at the earliest SNN time step. An explicit argmax would compare all $K$ similarities in one place, but each neuron holds only its own state. With TTFS, the comparison happens through spike timing instead: the winner is the prototype that fires first.
Unallocated prototypes (all synaptic weights zero) have $y^{(k)} = 0$ and thus $u^{(k)},v^{(k)} = 0$; they do not participate in TTFS competition. Both $v^{(k)}$ and $u^{(k)}$ are hard-reset to zero on threshold crossing or on receipt of the lateral-inhibition signal at DA3.
All prototype neurons connect all-to-all via DA3: any output spike resets $u^{(k)},v^{(k)} \leftarrow 0$ in every prototype, enforcing WTA competition.

The neuron emits a graded output spike with payload $\mathrm{id}^{(k)}$ (and also resets) when either of two conditions holds ($\alpha^{(k)}=1$ for all unallocated prototypes; see \hyperref[sec:methods_weight_update]{Weight update} below):
\begin{equation}\label{eq:spike_gen}
s^{(k)}(t) = 1 \;\iff\; \underbrace{v^{(k)}(t) \geq v_{\mathrm{th}}}_{\text{threshold crossing}} \;\;\lor\;\; \underbrace{\alpha^{(k)}(t)r^{(k)}(t) = 1}_{\text{modulator-forced spike}},
\end{equation}
where $s^{(k)}(t)\in\{0,1\}$ indicates whether prototype $k$ spikes at SNN time step $t$ (the spike carries the payload $\mathrm{id}^{(k)}$), and $r^{(k)}(t)$ is the modulatory variable that prototype $k$ reads from the signal $\phi(t)$ that the modulator neuron sends to the DA2 channels of all prototypes:
\begin{equation}\label{eq:modulator_interpretation}
r^{(k)}(t) = \begin{cases}
+1 & \text{if } \phi(t)=\mathrm{id}^{(k)}, \\
-1 & \text{if } \phi(t)= -\mathrm{id}^{(k)}, \\
0  & \text{otherwise.}
\end{cases}
\end{equation}
In other words, if the received modulator spike's payload matches $\mathrm{id}^{(k)}$, it is interpreted as positive reinforcement, while a payload of $-\mathrm{id}^{(k)}$ is negative reinforcement.

DA2 serves two roles: (i) any matched modulatory spike supplies the third factor of Eq.~(\ref{eq:loihi2_rule}), so that only the targeted prototype's synapses receive a non-zero update; and (ii) it forces an output spike in an unallocated prototype (Eq.~\ref{eq:spike_gen}). These two roles together implement novelty-triggered allocation: when no prototype fires within $t_{\mathrm{wait}}$ steps, the modulator targets the next free prototype $k_{\mathrm{new}}$, whose forced spike broadcasts a DA3 reset to all prototypes (restoring a clean state) and enables weight imprinting via the learning engine.

\textbf{Novelty-detector neuron.} The novelty-detector neuron has two dendritic accumulator compartments (two input channels) and two internal state variables: a countdown timer $\tau$ and an input-presence flag $m$. On arrival of any input-population spike at its first DA compartment, it sets:
\begin{equation}
m \leftarrow 1 \text{ and } \tau \leftarrow t_{\mathrm{wait}}.
\end{equation}
On each subsequent SNN time step while $m = 1$, the timer is decremented ($\tau \leftarrow \tau - 1$); the neuron fires a binary spike with payload $1$ (and resets $m, \tau \leftarrow 0$) when the countdown reaches zero ($\tau = 0$), i.e., once $t_{\mathrm{wait}}$ SNN time steps have elapsed without any prototype spike arriving on the second compartment. The second compartment acts as a canceling gate: a prototype spike arriving within $t_{\mathrm{wait}}$ suppresses the novelty signal. The coupling between prototype and novelty-detector parameters follows from noting that prototype $k$ spikes within the window iff $y^{(k)} + t_{\mathrm{wait}} \cdot \beta \geq v_{\mathrm{th}}$. Rearranging, the minimum feedforward activation required to suppress novelty detection is:
\begin{equation}\label{eq:novelty_threshold}
y_{\min} = v_{\mathrm{th}} - t_{\mathrm{wait}} \cdot \beta,
\end{equation}
which defines the \emph{effective similarity threshold}: inputs whose cosine similarity to every stored prototype falls below $y_{\min}$ trigger neurogenesis. This $y_{\min}$ is the threshold $\theta$ of the modulatory signal assignment below. Because $\beta$ and $t_{\mathrm{wait}}$ enter Eq.~(\ref{eq:novelty_threshold}) as a product, the two hyperparameters must be tuned jointly to achieve a desired novelty sensitivity.

\textbf{Modulator neuron.} The modulator neuron has two functions. As \emph{modulation relay}, it passes external supervisor feedback through: on receipt of a signed-identity signal $\pm\mathrm{id}^{(k^*)}$ from the supervisor (where the sign encodes correct vs.\ incorrect prediction), it emits a graded spike with that payload. At the same time, it acts as an \emph{allocator}: it maintains the index of the next unallocated prototype $\mathrm{id}_{\mathrm{next}}$; on receipt of a novelty-detector spike, it emits a graded spike with payload $\mathrm{id}_{\mathrm{next}}$ and increments $\mathrm{id}_{\mathrm{next}}$. Both signals reach the DA2 channels in the same format (Eq.~\ref{eq:modulator_interpretation}): the payload encodes the sign of the reward and the identity of the prototype to update. Allocation therefore needs no separate mechanism: it is the learning rule of Eq.~(\ref{eq:learning_rule_basic}) applied with $r^{(k)}=1$, $\alpha^{(k)}=1$ and $y^{(k)}=0$ (see \hyperref[sec:methods_weight_update]{Weight update} below).

\textbf{Winner-take-all competition and tie-break.} The winning prototype, which has the highest cosine similarity and fires first (Eq.~\ref{eq:ttfs}), broadcasts a DA3 reset. Its identifier, mapped to a class label by the supervisor, is the CLP-SNN prediction. When two or more prototypes cross threshold on the \emph{same} SNN time step, the chip emits multiple output spikes; these simultaneous winners are resolved off-chip by a host-side readout, which selects the winner by majority vote on the candidates' class labels with random tie-break among labels sharing the maximum vote count. The tie-break draw uses a fixed seed shared with the experiment-level random seed, so the entire learning trajectory is bit-reproducible across runs. When no prototype spikes, the class label of the input is assigned to the newly allocated prototype. Algorithm~\ref{alg:clp_snn} summarizes the full step: inference, winner selection and learning update.

\begin{algorithm*}[t]
\caption{CLP-SNN: one online continual-learning step for sample $(\mathbf{x}_n,\, \ell_n)$. On Loihi~2, Phase~3 is replaced by allocation of a new prototype after an incorrect prediction (Methods, ``Loihi 2 hardware simplification'').}
\label{alg:clp_snn}
\begin{algorithmic}[1]
\Require L2-normalized input $\mathbf{x}_n\!\in\!\mathbb{R}^d$, label $\ell_n$;\; prototype bank $\bigl\{(\mathbf{w}^{(k)},\,g^{(k)},\,\alpha^{(k)},\,\ell^{(k)})\bigr\}_{k=1}^{K}$;\; next free index $\mathrm{id}_{\mathrm{next}}$;\; parameters $v_{\mathrm{th}},\,\beta,\,t_{\mathrm{wait}},\,\Delta g$
\Statex
\Statex \textit{\textbf{Phase 1.} SNN inference by TTFS competition}
\State Input neurons fire at $t\!=\!0$; prototype $k$ accumulates $y^{(k)}\leftarrow\mathbf{w}^{(k)\top}\mathbf{x}_n$ via DA1 \Comment{dot product available at $t\!=\!1$}
\For{$t = 1,\ldots,t_{\mathrm{wait}}$} \Comment{all prototypes updated in parallel}
    \State $v^{(k)}(t) \leftarrow y^{(k)} + t\cdot\beta$ \Comment{bias ramp; active only for allocated prototypes ($y^{(k)}\!\neq\!0$)}
    \If{$v^{(k)}(t) \geq v_{\mathrm{th}}$}\; \textbf{emit spike}$(\mathrm{id}^{(k)})$ \Comment{TTFS: largest $y^{(k)}$ fires first}\EndIf
    \If{any prototype fired}\; \textbf{broadcast} DA3 reset $v^{(\cdot)}\leftarrow 0$;\; \textbf{break} \Comment{WTA lateral inhibition}\EndIf
\EndFor
\Statex
\Statex \textit{\textbf{Phase 2.} Winner selection}
\State $\mathcal{W}\leftarrow\{k : \text{prototype } k \text{ fired in Phase 1}\}$
\If{$\mathcal{W}=\{k'\}$}\; $k^*\leftarrow k'$
\ElsIf{$|\mathcal{W}|>1$} \Comment{simultaneous spikes: off-chip majority-vote tie-break}
    \State $c(\ell)\leftarrow\bigl|\{k\in\mathcal{W}:\ell^{(k)}=\ell\}\bigr|$;\; $\ell^*\leftarrow\operatorname{RandomChoice}\bigl(\arg\max_\ell c(\ell)\bigr)$
    \State $k^*\leftarrow\operatorname{RandomChoice}\!\bigl(\{k\in\mathcal{W}:\ell^{(k)}=\ell^*\}\bigr)$
\Else \Comment{$|\mathcal{W}|=0$, i.e.\ $\max_k y^{(k)}<\theta$: novelty detector fires}
    \State $k_{\mathrm{new}}\leftarrow\mathrm{id}_{\mathrm{next}}$;\; $\mathrm{id}_{\mathrm{next}}\leftarrow\mathrm{id}_{\mathrm{next}}+1$;\; $\mathbf{w}^{(k_{\mathrm{new}})}\leftarrow\mathbf{x}_n$;\; $\ell^{(k_{\mathrm{new}})}\leftarrow \ell_n$ \Comment{neurogenesis (Eq.~\ref{eq:learning_rule_basic}, $r\!=\!\alpha\!=\!1$, $\mathbf{w}\!=\!\mathbf{0}$)}
    \State \Return
\EndIf
\Statex
\Statex \textit{\textbf{Phase 3.} Supervised learning update (winner $k^*$)}
\State $r^{(k^*)}\leftarrow +1$ if $\ell^{(k^*)}=\ell_n$, else $-1$ \Comment{reward signal from supervisor}
\State $g^{(k^*)}\leftarrow\max\!\bigl(1,\,g^{(k^*)}+\Delta g\,r^{(k^*)}\bigr)$;\quad $\alpha^{(k^*)}\leftarrow 1/g^{(k^*)}$ \Comment{metaplasticity (Eqs.~\ref{eq:goodness_trace}--\ref{eq:alpha_trace})}
\State $\mathbf{w}^{(k^*)}\leftarrow\mathbf{w}^{(k^*)}+\alpha^{(k^*)}\cdot\bigl(r^{(k^*)}\mathbf{x}_n-|r^{(k^*)}|\,y^{(k^*)}\mathbf{w}^{(k^*)}\bigr)$ \Comment{self-normalizing learning rule (Eq.~\ref{eq:learning_rule_basic})}
\end{algorithmic}
\end{algorithm*}

\textbf{Modulatory signal assignment.} For each sample, the third-factor signal $r_n^{(k)}$ that prototype $k$ receives is:

\begin{align}
r_n^{(k)} = \begin{cases}
+1 & \text{if } k = k^* \text{ and prediction is correct} \\
-1 & \text{if } k = k^* \text{ and prediction is incorrect} \\
+1 & \text{if $\max_{k'} y_{n}^{(k')} < \theta$ and $k = k_{\mathrm{new}}$} \\
0 & \text{otherwise}
\end{cases}
\end{align}

\phantomsection\label{sec:methods_weight_update}\textbf{Weight update.} Each prototype neuron's adaptive learning rate $\alpha^{(k)}$ scales its weight update; weights are initialized to zero and $\alpha^{(k)}=1$. At allocation, Eq.~(\ref{eq:learning_rule_basic}) with $\alpha^{(k)}=1$ imprints the normalized input directly onto the weights, because the decay term vanishes for zero weights. Subsequent updates use the $\alpha^{(k)}$ set by the goodness score (Supplementary Figs.~\ref{fig:learning_rule_input_d_sensitivity},~\ref{fig:learning_rule_sigma_analysis}).

\textbf{Learning-rate update.} Following CLP~\cite{hajizada2024clp}, the winning neuron's learning rate is adjusted via a \emph{goodness} score $g$ (incremented by $\Delta g$ on correct predictions and decremented by $\Delta g$ on incorrect ones, floored at 1; Eqs.~\ref{eq:goodness_trace}--\ref{eq:alpha_trace}, Supplementary Table~\ref{tab:hyperparameters}), with correctness determined by supervisor label-matching. The next section describes how these neuron models are deployed on Loihi~2.

\subsection*{Loihi 2 deployment}
\phantomsection\label{sec:methods_deployment}

Intel Loihi~2 is a digital neuromorphic chip supporting up to one million neurons and 120 million synapses across 128 neurocores~\cite{intel2021loihi2brief,orchard2021loihi,shrestha2024efficient}. It operates with asynchronous communication within each time step and barrier synchronization between time steps, yielding variable-length SNN time steps (minimum 200~ns) that scale with computational load. Each neurocore supports up to 8,192 neurons with microcode-programmable neuron models and dense, sparse, or convolutional connectivity. It provides 32~KB of neuron SRAM plus up to 128~KB of synapse SRAM, which can be flexibly partitioned between neuron state and synaptic storage.

The four CLP-SNN neuron populations specified above map onto Loihi~2 as follows. The input population, a stateless graded-spike relay layer with one neuron per feature dimension, fits on a single core. The prototype population occupies the majority of the allocated cores; at the network size used for benchmarking (1280 input dimensions, up to 300 prototypes), CLP-SNN fits within 41 of the 128 neurocores of a single chip. Lava's partitioner places each neuron population on its own cores, so the novelty-detector and modulator neurons occupy one core each. Supplementary Fig.~\ref{fig:loihi_deployment} shows the concrete core assignment produced by the Lava partitioner.

Each chip-level primitive that CLP-SNN depends on is anchored to a specific neuron mechanism defined in the \hyperref[sec:methods_architecture]{Architecture subsection} above. The microcode-programmable neuron primitive of Loihi~2 is what enables the custom prototype, novelty-detector and modulator dynamics; on previous-generation Loihi, which exposes only hard-coded leaky-integrate-and-fire neurons, these mechanisms cannot be implemented natively. The integer-valued (graded) spike primitive is what allows the prototype neuron's feedforward channel (DA1) to accumulate the full dot product $y^{(k)} = \sum_j w^{(k)}_j x_j$ within a single SNN time step. The flexible dendritic-compartment primitive routes feedforward input spikes, third-factor modulatory spikes and lateral-inhibition spikes to the three input channels (DA1, DA2, DA3) of each prototype neuron independently, so that each payload enters the neuron program as a distinct variable: the feedforward activation and the third factor feed the learning rule (Eq.~\ref{eq:loihi2_rule}), and the lateral-inhibition spike triggers the WTA reset. Finally, the input-dependent overwrite of pre-synaptic traces (Eq.~\ref{eq:pre_traces}) makes the graded input values directly available to the learning engine as pre-synaptic variables. The CLP-SNN deployment on Loihi~2 is functionally equivalent to the architecture of Fig.~\ref{fig:figure_2}, with one exception: each prototype is imprinted once at allocation and not updated afterwards, and a misclassified input allocates a new prototype (see \hyperref[sec:methods_hw_simplification]{``Loihi 2 hardware simplification''} below).

\textbf{Supervisor.} The supervisor is a lightweight CPU-side process that performs three operations per sample: (1) a lookup of the winner's class label in a prototype-index-to-class-label mapping table, (2) a scalar integer comparison between this predicted label and the ground-truth class label, and (3) transmission of the result to the chip, as the signed-identity feedback $\pm\mathrm{id}^{(k^*)}$ of the full rule or, in the imprint-only deployment, as a trigger to the modulator to allocate a new prototype when the prediction is wrong. This exchange between chip and host is IO and, like all IO, is excluded from the 0.33~ms per-sample latency.

\subsection*{On-chip implementation of the learning rule} 
Loihi~2 supports programmable sum-of-product learning rules acting on synaptic variables and activity traces~\cite{davies2018loihiLearning}. Each synapse has three programmable variables (Weight, Delay, Tag) and access to two pre-synaptic and three post-synaptic traces. On previous-generation Loihi, these traces are exponentially filtered spike trains. Loihi~2 also lets a pre-synaptic trace be overwritten with the payload of an incoming graded spike, and lets the neuron write its post-synaptic traces directly. These two mechanisms enable the input-dependent trace assignment of Eqs.~(\ref{eq:pre_traces}--\ref{eq:post_traces}). Products in the learning rule can depend on pre- and post-synaptic variables, and the learning engine updates the weights in parallel across cores once per learning epoch, a programmable number of SNN time steps ($t_{\mathrm{epoch}} = 20$ here, one sample). Near-memory computing, with the weight-update logic co-located with the synaptic SRAM, makes on-chip learning fast and energy-efficient.

The learning rule of Eq.~(\ref{eq:learning_rule_basic}) maps onto the Loihi~2 learning engine as follows:
\begin{equation} \label{eq:loihi2_rule}
\Delta w_{ij}(t) = \alpha_i(t)\, s_i^r(t)\, x_j(t) - \alpha_i(t)\, \bigl|s_i^r(t)\bigr|\, w_{ij}(t)\, y_i(t),
\end{equation}
where $\Delta w_{ij}(t)$ is the weight change of the synapse from pre-synaptic (input) neuron $j$ to post-synaptic (prototype) neuron $i$ at SNN time step $t$, $\alpha_i(t)$ is the adaptive learning rate, $s_i^r(t)$ is the third-factor (reward) spike train that neuron $i$ reads from the modulatory signal $\phi(t)$ (Eq.~\ref{eq:modulator_interpretation}), $x_j(t)$ is the pre-synaptic trace and $y_i(t)$ is the post-synaptic trace. Eq.~(\ref{eq:loihi2_rule}) is Eq.~(\ref{eq:learning_rule_basic}) written per synapse and per SNN time step. With $i=k$, $\Delta w_{ij}$ and $x_j$ are the $j$-th components of the update of $\mathbf{w}^{(k)}$ and of $\mathbf{x}_n$, and $\alpha_i$, $s_i^r$ and $y_i$ correspond to $\alpha^{(k)}$, $r^{(k)}$ and $y^{(k)}$. Their evolution within a sample follows Eq.~(\ref{eq:modulator_interpretation}) and Eqs.~(\ref{eq:pre_traces}--\ref{eq:alpha_trace}). The prototype neuron holds the third factor from its arrival until the end of the learning epoch, when the learning engine applies Eq.~(\ref{eq:loihi2_rule}). The spike trains are written as

\begin{equation} \label{eq:spike_trains}
s_i^m(t) = \sum_f a_i^{m,f}\,\delta(t-t_i^{m,f}),
\end{equation}
where $t_i^{m,f}$ is the time of the $f$-th spike of train $m \in \{x,r\}$ (pre-synaptic or third-factor) of neuron $i$, $a_i^{m,f}$ is its graded amplitude, and $\delta$ is the Kronecker delta. Pre- and post-synaptic traces update as:

\begin{equation} \label{eq:pre_traces}
x_j(t+1) = \begin{cases}
  s_j^x(t) & \text{if input neuron } j \text{ spikes at } t, \\
  x_j(t) & \text{otherwise.}
\end{cases}
\end{equation}

\begin{equation} \label{eq:post_traces}
y_i(t+1) = \begin{cases}
  \sum_j w_{ij}(t)\, x_j(t+1) & \text{if input spikes arrive}, \\
  y_i(t) & \text{otherwise.}
\end{cases}
\end{equation}
The traces are overwritten when a sample is injected and held until the next one. Unlike $x_j$, $y_i$ is not derived from a spike train: the prototype neuron writes it from its DA1 accumulation. Learning rate $\alpha_i$ and goodness $g_i$ update on arrival of supervisor feedback (Eqs.~\ref{eq:goodness_trace}--\ref{eq:alpha_trace}).

\begin{equation} \label{eq:goodness_trace}
g_i(t+1) = \begin{cases}
  \max(1,\, g_i(t)+\Delta g\, s_i^r(t)) & \text{if } s_i^r(t) \neq 0, \\
  g_i(t) & \text{if } s_i^r(t) = 0.
\end{cases}
\end{equation}

\begin{equation} \label{eq:alpha_trace}
\alpha_i(t+1) = \begin{cases}
  1/g_i(t+1) & \text{if } s_i^r(t) \neq 0, \\
  \alpha_i(t) & \text{if } s_i^r(t) = 0.
\end{cases}
\end{equation}

\phantomsection\label{sec:methods_hw_simplification}\textbf{Loihi 2 hardware simplification.} The on-chip CLP-SNN deployment operates under a simplified version of the learning rule described above. On Loihi~2, each prototype is imprinted \emph{exactly once}: at allocation time the learning engine executes Eq.~(\ref{eq:loihi2_rule}) with $\alpha_i=1$, $s_i^r=+1$ and $y_i=0$, which reduces to $\Delta w_{ij} = x_j$, directly imprinting the input feature vector onto the prototype weights. After this single update the weights are frozen; no further learning occurs at that synapse regardless of subsequent predictions. When an allocated prototype makes an incorrect prediction, the Loihi~2 system allocates a \emph{new} prototype for the input rather than applying a corrective update (the $s_i^r=-1$ case of Eq.~\ref{eq:loihi2_rule}), making the on-chip path equivalent to the novelty-allocation path for both novel and misclassified inputs. Consequently, the goodness and adaptive learning-rate dynamics of Eqs.~(\ref{eq:goodness_trace}--\ref{eq:alpha_trace}) are not active on hardware. Current software limitations restrict the on-chip deployment to imprinting. Imprint-only operation needs about twice as many prototypes as the fully adaptive CLP-SNN (\hyperref[supp:adaptive]{Supplementary Note~2}), but each weight update costs the same in both modes, because the learning engine sweeps every plastic synapse at each learning epoch (Supplementary Table~\ref{tab:op_counts}). The full adaptive rule of this section runs in simulation, in floating point and in an 8-bit integer (INT8) variant that follows the arithmetic of the Loihi~2 learning engine and is compatible with the chip. In this variant, weights and inputs are integers in $[-128,127]$ at scale $S=128$ ($x_{\mathrm{int}} = \mathrm{round}(x\cdot S)$, saturating clamp on overflow). Each weight update is computed in a 15-bit signed accumulator and stochastically rounded to 8 bits with a single shared scalar draw per update, matching the Lava bit-approximate hardware spec.

\subsection*{Experimental framework}
\textbf{Dataset and preprocessing.} For all experiments, we used the OpenLORIS dataset~\cite{she2020openloris}. The dataset includes 36 video clips for each of 121 objects, which are divided into 40 object categories. We used the category-level labels. We took the first 60 frames from each video clip to remove the imbalance between classes, as video clips have variable lengths. We adopted the off-the-shelf ImageNet-trained EfficientNet-B0 model~\cite{tan2019efficientnet} as the static feature extractor, which is widely used for edge devices~\cite{hayes2022online}. It runs as a static, pre-trained model on a conventional GPU of the host computer. Only the 1280-dimensional feature vectors extracted by this backbone are clipped to non-negative values, L2-normalized and passed to Loihi 2 as inputs to the CLP-SNN continual learning layer. This frozen-backbone protocol is the standard one for OCL benchmarking~\cite{hayes2022online,hajizada2024clp} and isolates the OCL algorithm under test from feature-space adaptation. Adapting the feature extractor continually, deploying it on-chip and comparing feature extractors are beyond the scope of this paper.

For both 1-shot and 25-shot learning, each shot corresponds to a collection of one random video clip for each of 40 object classes. For both setups, we ran three experiments with different random seeds, each randomizing the class order and the choice of video clips. All reported quantities are mean $\pm$ standard deviation across these three independent runs.

\phantomsection\label{sec:methods_baselines}\textbf{Baseline methods.} We employed Streaming Linear Discriminant Analysis (SLDA), Nearest Class Mean (NCM), Replay, Fine-tuning and Perceptron as baseline methods, following the evaluation protocol of~\cite{hayes2022online,hajizada2024clp}. SLDA~\cite{hayes2020deepSLDA} classifies via a Mahalanobis-distance decision rule using a shared streaming precision matrix. NCM maintains one running-average prototype per class and assigns each test sample to the closest class centroid; on L2-normalized features, cosine and Euclidean distance are equivalent and either can be used. Replay keeps a class-balanced buffer of 800 past samples and replays 50 of them, drawn at random, alongside each new sample to mitigate catastrophic forgetting. Fine-tuning and Perceptron are non-continual baselines that update last-layer weights online without any forgetting-mitigation mechanism, and serve here to quantify the accuracy cost of catastrophic forgetting. No neuromorphic comparator exists under this protocol: neuromorphic implementations are chip-specific, and the Loihi 1 systems cited in the \hyperref[sec:intro]{Introduction} have no Loihi 2 implementation.

Our implementations start from the Embedded-CL repository accompanying~\cite{hayes2022online} (\url{https://github.com/tyler-hayes/Embedded-CL}). The only modifications are L2 normalization of the inputs, the exact rank-one SLDA update of \hyperref[sec:methods_rankone_slda]{``Exact rank-one streaming LDA baseline''}, and data-path optimizations that leave the computed result identical. All methods except Replay receive L2-normalized features. For SLDA this departs from the reference implementation, which does not normalize; Replay is left unnormalized because its cross-entropy-based fine-tuning objective is poorly conditioned on the unit hypersphere, which degrades its performance under normalization.

\subsection*{Exact rank-one streaming LDA baseline}
\phantomsection\label{sec:methods_rankone_slda}
Streaming LDA~\cite{hayes2020deepSLDA} classifies a feature vector $\mathbf{x}_n \in \mathbb{R}^{d}$ into one of $C$ classes. It keeps the mean $\boldsymbol{\mu}_\ell$ of the $n_\ell$ samples seen so far from each class $\ell$, and a single covariance $\boldsymbol{\Sigma}_n$ shared by all classes. A new sample $\mathbf{x}_n$ with label $\ell_n$ updates the covariance as
\begin{equation}
\boldsymbol{\Sigma}_{n+1} = \frac{n\boldsymbol{\Sigma}_n + c_n\,\mathbf{z}_n\mathbf{z}_n^{\top}}{n+1},
\quad c_n = \frac{n}{n+1},
\quad \mathbf{z}_n = \mathbf{x}_n - \boldsymbol{\mu}_{\ell_n},
\label{eq:slda_cov}
\end{equation}
where $n$ is the number of samples processed before $\mathbf{x}_n$ and $\mathbf{z}_n$ is the deviation of the sample from its class mean before that mean is updated. Classification uses a precision matrix $\boldsymbol{\Lambda}$, a regularized inverse of the covariance. A sample $\mathbf{x}$ is assigned to the class with the highest linear score $\mathbf{x}^{\top}\boldsymbol{\omega}_\ell - b_\ell$, with discriminant weights $\boldsymbol{\omega}_\ell = \boldsymbol{\Lambda}\boldsymbol{\mu}_\ell$ and bias $b_\ell = \tfrac{1}{2}\boldsymbol{\mu}_\ell^{\top}\boldsymbol{\omega}_\ell$. In matrix form, $\boldsymbol{\Omega} = \boldsymbol{\Lambda}\mathbf{M}$, where $\mathbf{M} \in \mathbb{R}^{d\times C}$ stacks the class means as columns; a subscript $n$ marks values after $n$ samples. The reference implementation~\cite{hayes2020deepSLDA,hayes2022online} recomputes this inverse whenever the statistics have changed since the last prediction. In the batch-size-1 protocol this happens once per sample, at a cost of $O(d^{3})$.

To avoid this inversion, we update the inverse directly instead of recomputing it. This requires the matrix being inverted to change by a single outer product per sample, which takes two changes of form: the recursion is written for the unnormalized scatter, and the covariance shrinkage is replaced by a fixed ridge. For the scatter $\mathbf{S}_n = n\boldsymbol{\Sigma}_n$, which removes the normalization by $n+1$, Eq.~(\ref{eq:slda_cov}) reads $\mathbf{S}_{n+1} = \mathbf{S}_n + c_n\mathbf{z}_n\mathbf{z}_n^{\top}$. Regularizing the scatter with a fixed ridge $\lambda > 0$ keeps this update rank-one:
\begin{equation}
\mathbf{A}_{n+1} = \mathbf{A}_n + c_n\,\mathbf{z}_n\mathbf{z}_n^{\top},
\quad \mathbf{A}_n = \mathbf{S}_n + \lambda\mathbf{I}.
\label{eq:slda_rank1}
\end{equation}
Because each sample changes $\mathbf{A}_n$ by a rank-one term, its inverse $\boldsymbol{\Lambda}_n = \mathbf{A}_n^{-1}$, which serves as the precision matrix, also changes by a rank-one term. The classifier is therefore updated in three steps, none of which inverts a matrix. First, the Sherman--Morrison identity~\cite{sherman1950adjustment,bartlett1951inverse} gives the new precision exactly, at a cost of $O(d^{2})$:
\begin{equation}
\begin{gathered}
\boldsymbol{\Lambda}_{n+1} = \boldsymbol{\Lambda}_n - \gamma_n\,\mathbf{h}_n\mathbf{h}_n^{\top},\\
\mathbf{h}_n = \boldsymbol{\Lambda}_n\mathbf{z}_n,
\quad q_n = \mathbf{z}_n^{\top}\mathbf{h}_n,
\quad \gamma_n = \frac{c_n}{1 + c_n q_n}.
\end{gathered}
\label{eq:slda_sm}
\end{equation}
Here $\mathbf{h}_n$ is the deviation $\mathbf{z}_n$ transformed by the current precision, so the patch is a single outer product scaled by a scalar. Second, the discriminant weights inherit the same patch. Multiplying Eq.~(\ref{eq:slda_sm}) on the right by $\mathbf{M}_n$ gives the weights under the new precision and the old means, at a cost of $O(dC)$:
\begin{equation}
\boldsymbol{\Lambda}_{n+1}\mathbf{M}_n = \boldsymbol{\Omega}_n - \gamma_n\,\mathbf{h}_n\mathbf{s}_n^{\top},
\quad \mathbf{s}_n = \mathbf{M}_n^{\top}\mathbf{h}_n.
\label{eq:slda_wpatch}
\end{equation}
Third, the mean of class $\ell = \ell_n$ moves by $\mathbf{z}_n/(n_\ell+1)$. The new weights $\boldsymbol{\Omega}_{n+1} = \boldsymbol{\Lambda}_{n+1}\mathbf{M}_{n+1}$ therefore differ from Eq.~(\ref{eq:slda_wpatch}) only in column $\ell$, by $\boldsymbol{\Lambda}_{n+1}\mathbf{z}_n/(n_\ell+1)$. This needs no second matrix--vector product: by Eq.~(\ref{eq:slda_sm}), $\boldsymbol{\Lambda}_{n+1}\mathbf{z}_n = \mathbf{h}_n(1-\gamma_n q_n) = \mathbf{h}_n/(1+c_n q_n)$, which has the form of the gain vector of recursive least squares~\cite{plackett1950theorems,haykin2002adaptive}. Hence
\begin{equation}
\begin{aligned}
\boldsymbol{\mu}_{\ell} &\leftarrow \boldsymbol{\mu}_{\ell} + \frac{\mathbf{z}_n}{n_\ell + 1},\\
\boldsymbol{\omega}_{\ell} &\leftarrow \boldsymbol{\omega}_{\ell} + \frac{\mathbf{h}_n}{(1 + c_n q_n)(n_\ell + 1)},
\end{aligned}
\qquad \ell = \ell_n.
\label{eq:slda_meanshift}
\end{equation}
The biases $b_\ell$ are updated alongside, so a prediction needs a single vector--matrix product, $\hat{\ell} = \arg\max_{\ell}\left(\mathbf{x}^{\top}\boldsymbol{\omega}_\ell - b_\ell\right)$ over the classes seen so far. The update costs $2d^{2} + 2dC \approx 3.4\times10^{6}$ multiply--accumulate operations per sample at $d = 1280$ and $C = 40$, against roughly $2\times10^{9}$ for one explicit inversion. Starting from $\boldsymbol{\Lambda}_0 = \mathbf{I}/\lambda$, Eq.~(\ref{eq:slda_sm}) gives $\boldsymbol{\Lambda}_n = (\mathbf{S}_n + \lambda\mathbf{I})^{-1}$ exactly for every $n$. The maintained classifier is therefore identical, up to floating-point rounding, to one that recomputes this inverse at every sample (\hyperref[supp:rankone]{Supplementary Note~3}). Because $\mathbf{A}_n$ is positive definite, $q_n \geq 0$ and $1 + c_n q_n \geq 1$ for all $n$, so the update never divides by a small number.

The reference implementation instead regularizes by shrinking the covariance toward the identity, which is not a rank-one change. The fixed ridge $\lambda$ of Eq.~(\ref{eq:slda_rank1}) differs from this shrinkage only in its schedule. We selected $\lambda = 1$ once on a held-out class order. With this value, the rank-one implementation is at least as accurate as the reference implementation on both protocols (Supplementary Fig.~\ref{fig:ea_equivalence}). \hyperref[supp:rankone]{Supplementary Note~3} gives the regularization equivalence and the numerical verification.

Rank-one maintenance of a regularized inverse is a classical device. Bartlett~\cite{bartlett1951inverse} introduced it for discriminant analysis, and it is routine in recursive least squares~\cite{plackett1950theorems,haykin2002adaptive}. It has been applied to online linear discriminant classifiers under concept drift~\cite{kuncheva2008adaptive,yeh2013rankone} and to fixed-update-time incremental classification of robot-collected objects~\cite{camoriano2017incremental}, and it underlies analytic continual learning~\cite{zhuang2022acil,mcdonnell2023ranpac}. To our knowledge it has not been applied to the streaming LDA baseline~\cite{hayes2020deepSLDA,hayes2022online} in online continual-learning benchmarking. We release the implementation together with its exactness and parity tests.

\subsection*{Cross-platform latency and energy benchmarking}
\phantomsection\label{sec:methods_benchmarking}
The comparison across platforms uses the same feature vectors from the same off-the-shelf model for all methods, identical evaluation protocols and metrics, and the same random seeds. All baseline methods were benchmarked for energy consumption and latency on an NVIDIA Jetson Orin Nano (8~GB, 15~W TDP) running JetPack 6.2.1 with PyTorch 2.4.0. This platform was selected as a representative AI edge device comparable to Loihi for on-device computing applications. Energy measurements captured total system power consumption (VDD\_IN) using the Tegrastats monitoring tool. Each baseline was run on both the GPU and the CPU of the Orin. CLP-SNN was evaluated on one N3C2-revision Loihi~2 chip of an Oheo Gulch system, using Lava 0.11.0.dev0, Lava-Loihi 0.6.0.dev0 and NxCore 2.5.8.

On both platforms, dynamic energy is the total energy minus the static energy over the same time; static power is the mean VDD\_IN over 30~s of idle on the Orin and was measured with no network running on Loihi~2. For consistency across all platforms, IO latencies were excluded from GPU, CPU and Loihi 2 benchmark measurements. Each method is charged for one test-then-train step per sample: the prediction for the incoming sample followed by the update on its label (Table~\ref{table:1}). On the Orin, the two primitives of this step, the query (scoring the incoming sample) and the update (learning from its label), were timed separately over the 1-shot stream after a warm-up pass, using CUDA events on the GPU and the wall clock on the CPU. For CLP and Replay, the update already computes the prediction (winner selection and the training forward pass, respectively), so the step is charged as the update alone. For NCM and SLDA, the update does not score the sample, so the step is charged as the sum of query and update. Supplementary Table~\ref{tab:cost_primitives} gives the value of each primitive.

We considered INT8 GPU inference as an additional baseline to provide a stricter precision-matched comparison with Loihi~2's INT8 arithmetic. On Jetson Orin Nano, INT8 acceleration is exposed through tensor cores, which support only dense General Matrix Multiply (GEMM) operations at large batch sizes. PyTorch's autograd does not support INT8 gradients; the dominant operations of the learning step (rank-one covariance and precision patches, element-wise prototype updates) are not GEMM, and the OCL protocol operates at a batch size of 1. Consequently, INT8 acceleration on the Orin Nano GPU is infeasible for the learning step of the evaluated methods. We instead bracket the GPU baseline with FP32 (standard) and FP16 (half precision) runs (Supplementary Table~\ref{tab:cost_primitives}); CLP-SNN's latency and energy advantages persist against both.

We performed all characterization experiments with an input dimensionality of 1280, which is the penultimate layer size of the EfficientNet-B0 backbone. The number of classes is set to 40 for all methods. The number of prototypes pre-allocated for both CLP and CLP-SNN is set to 300. The corresponding on-chip footprint is reported in the \hyperref[sec:methods_deployment]{Loihi 2 deployment} subsection of Methods. The complete set of CLP-SNN hyperparameters used throughout this work is listed in Supplementary Table~\ref{tab:hyperparameters}. Per-sample operation counts for every learner at this working point, and the inventory of synaptic and plasticity operations that the Lava-Loihi compiler instantiates for CLP-SNN, are given in Supplementary Table~\ref{tab:op_counts}.

\section*{Data availability}
The OpenLORIS object recognition dataset used in this study is publicly available from its original authors~\cite{she2020openloris} (\url{https://lifelong-robotic-vision.github.io/dataset/object}). Pre-trained EfficientNet-B0 ImageNet weights~\cite{tan2019efficientnet} were used as released through standard public deep-learning libraries. The extracted EfficientNet-B0 feature vectors are deposited on Zenodo (\url{https://doi.org/10.5281/zenodo.20492557}). The accuracy/latency/energy benchmarking outputs and figure-source data files generated in this study are deposited in the public reproducibility repository accompanying this paper~\cite{clpsnn_code}. Raw Loihi~2 measurement logs and Intel proprietary software internals are not redistributable; aggregated per-sample latency and energy measurements derived from those logs are included in the reproducibility repository.

\section*{Code availability}
The simulation code for CLP-SNN (floating-point and INT8 variants), the convergence analysis of the Taylor-expansion self-normalizing rule, 1-shot and 25-shot OCL experiments, forgetting analysis, Orin benchmarking scripts and figure-generation notebooks are openly available at \url{https://github.com/elvinhajizada/CLP-SNN}~\cite{clpsnn_code}. The Loihi~2 hardware implementation cannot be openly redistributed because it depends on Intel's proprietary Lava-Loihi software stack and Loihi~2 hardware, both of which are governed by the Intel Neuromorphic Research Community (INRC) license; the source code is available to INRC members and other interested researchers may contact the authors to arrange collaboration through the INRC.

\bibliographystyle{naturemag}
\bibliography{references}  %

\section*{Acknowledgments}
This work was performed at Intel Labs, Neuromorphic Computing Lab, and was funded by Intel Corporation. E.H. additionally acknowledges support by the German Research Foundation (Deutsche Forschungsgemeinschaft, DFG; project number 534429653).

\appendix  %
\onecolumn

\setcounter{table}{0}
\setcounter{figure}{0}
\renewcommand{\thetable}{S\arabic{table}}  %
\renewcommand{\thefigure}{S\arabic{figure}} %
\renewcommand{\thealgorithm}{S\arabic{algorithm}} %
\setcounter{algorithm}{0}

\section*{Supplementary Material}

\subsection*{Supplementary Figures}

\begin{figure*}[h!]
    \centering
    \includegraphics[width=0.995\textwidth]{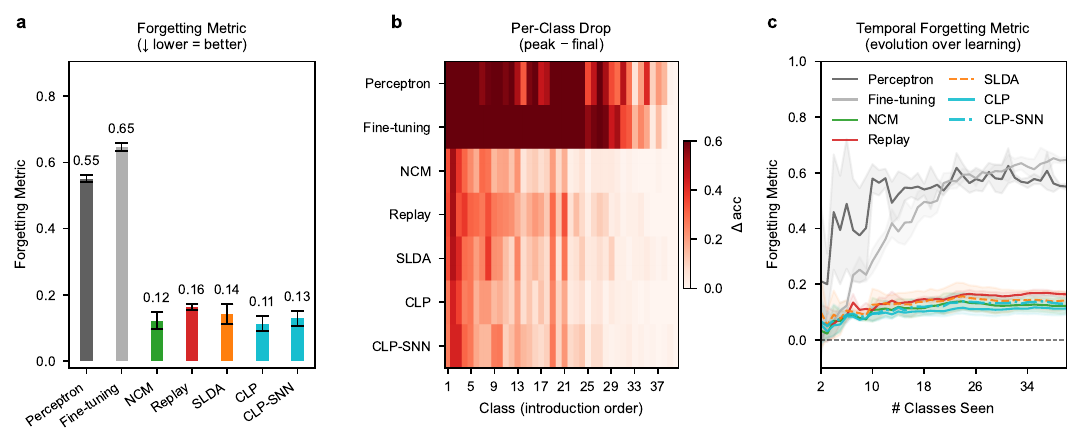}
    \caption{\textbf{Catastrophic forgetting analysis for all OCL methods under the 1-shot class-incremental protocol.} \textbf{a} Forgetting Metric (FM) for each method, averaged over all classes and three random seeds (mean $\pm$ std). Lower is better. FM~\cite{chaudhry2018riemannianWalk} is defined as the drop from a class's peak accuracy to its final accuracy, FM$_j = \max_t R[t,j] - R[N-1,j]$ (where $R[t,j]$ is the test accuracy on class $j$ after training through step $t$, and $N = 40$ is the number of steps), and is averaged across all classes $j$. This forgetting measure confounds actual loss of knowledge with an accuracy drop due to the classification task becoming harder as new classes are added over the course of learning. Nevertheless, it is the practically relevant measure in any realistic class-incremental continual learning setting. Non-continual baselines (Perceptron, Fine-tuning) show severe forgetting (FM $\geq 0.55$); OCL methods cluster between $0.11$ and $0.16$. \textbf{b} Per-class forgetting heatmap: each cell shows FM$_j$ for one method (row) and one class (column, ordered by introduction sequence). Dark cells indicate classes that have been forgotten more. All methods' forgetting concentrates on early-introduced classes, as expected. However, non-continual baselines exhibit very large forgetting across almost two-thirds of the classes, while retaining knowledge of the last few classes. The prototype-based OCL methods (NCM, CLP and CLP-SNN) show less forgetting and more uniformly light cells, indicating stable retention across all class positions. Replay shows more forgetting than the prototype-based methods. \textbf{c} Temporal evolution of the forgetting metric as the number of seen classes grows. Shaded bands show $\pm 1$ standard deviation across seeds. Perceptron and Fine-tuning accumulate forgetting rapidly; CLP and CLP-SNN stabilize at low forgetting and track each other closely.}
    \label{fig:detailed_forgetting}
\end{figure*}

\begin{figure*}[h!]
    \centering
    \includegraphics[width=0.999\textwidth]{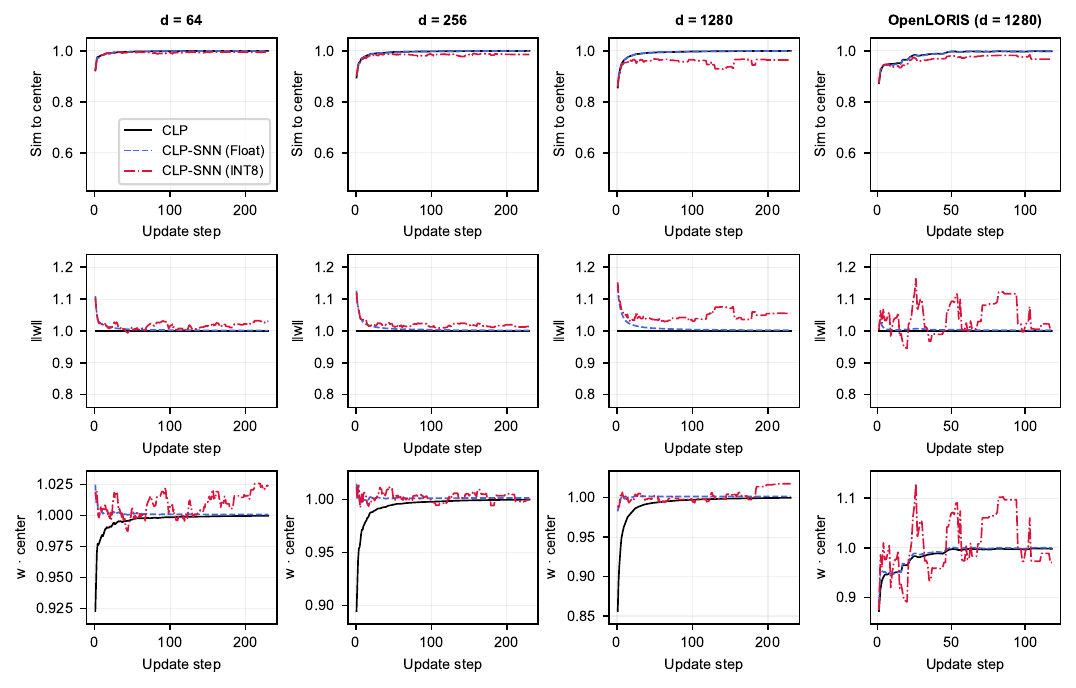}
    \caption{\textbf{Learning rule convergence and weight-norm stability across input dimensionalities.} The first three columns show synthetic clusters at dimensionality $d \in \{64, 256, 1280\}$, and the rightmost column shows a real OpenLORIS class (class 13). Synthetic clusters are generated with a fixed effective spread: the minimum cosine similarity of any cluster point to its center is held at $\geq 0.75$, matching CLP's allocation threshold; the same criterion selects 119 of the 180 training frames of the OpenLORIS class. Three rows track three complementary diagnostics over 230 update steps (118 for the OpenLORIS class): \textbf{(Row 1)} cosine similarity of the prototype weight vector $\mathbf{w}$ to the true cluster center; \textbf{(Row 2)} L2 norm of $\mathbf{w}$ (ideal $= 1$); \textbf{(Row 3)} raw (unnormalized) dot product $\mathbf{w} \cdot \mathbf{c}_{\text{center}}$, which conflates alignment and norm growth. Three variants are compared: CLP (explicit L2 renormalization after each update, black), CLP-SNN Float (Taylor self-normalizing rule, 64-bit, blue) and CLP-SNN INT8 (integer arithmetic of the Loihi~2 learning engine, red dashed). CLP and CLP-SNN Float converge nearly identically across all dimensionalities. The initially higher norm of CLP-SNN Float results from the high initial learning rates of CLP's learning-rate adaptation, which violate the small-learning-rate assumption of the Taylor expansion. The norm therefore overshoots at first and converges to one as the learning rates decay. CLP-SNN INT8 tracks both variants closely at low dimensionality. At high dimensionality the feature values per dimension are smaller, so once the learning rate has decayed many weight updates fall below one step of the 8-bit weight, and the INT8 prototype settles at a higher norm and a lower cosine similarity, with 5--10\% approximation errors. The two errors offset each other in the dot product, the similarity measure the network uses, which converges to one. The OpenLORIS column shows the same behavior on real features.}
    \label{fig:learning_rule_input_d_sensitivity}
\end{figure*}

\begin{figure*}[h!]
    \centering
    \includegraphics[width=0.995\textwidth]{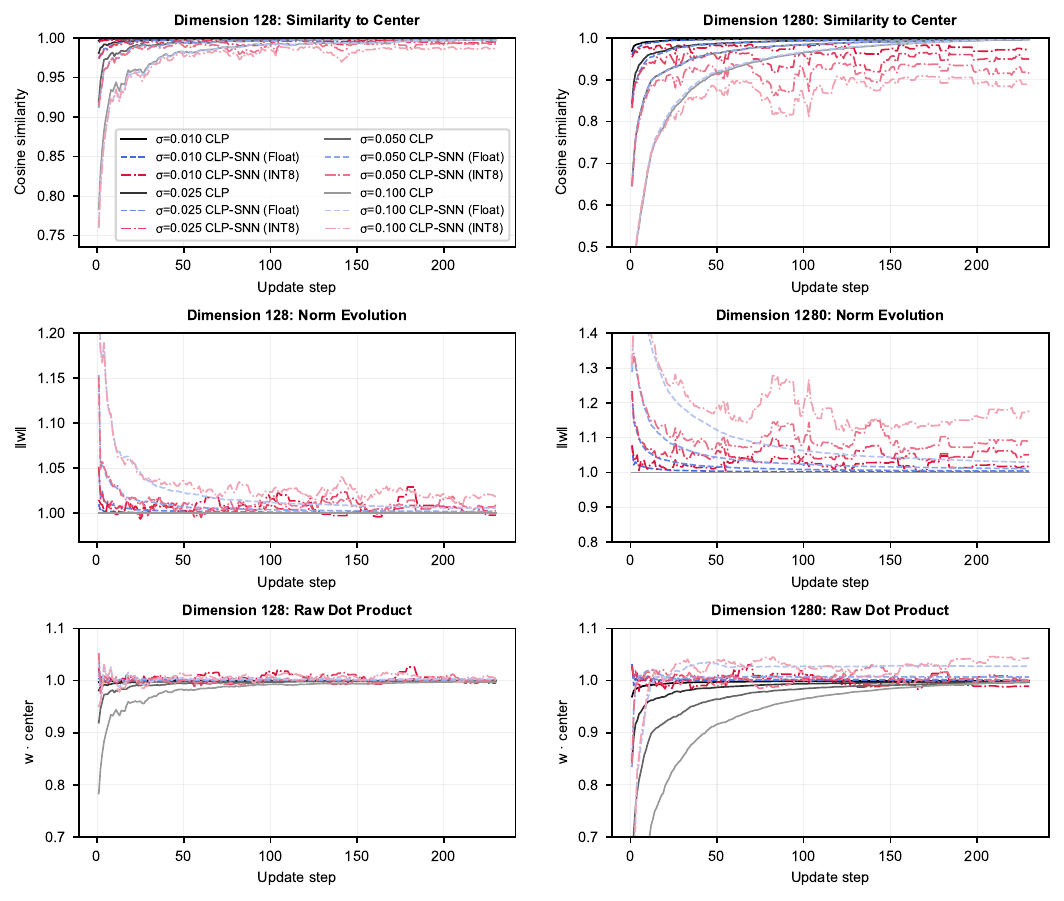}
    \caption{\textbf{Sensitivity of the self-normalizing learning rule to intra-class cluster spread $\sigma$.} Synthetic clusters are generated at two representative dimensionalities ($d = 128$, left column; $d = 1280$, right column) with isotropic Gaussian spread at four levels $\sigma \in \{0.010, 0.025, 0.050, 0.100\}$. Three rows show the same diagnostics as Supplementary Fig.~\ref{fig:learning_rule_input_d_sensitivity}: \textbf{(Row 1)} cosine similarity to the cluster center; \textbf{(Row 2)} L2 norm evolution; \textbf{(Row 3)} raw dot product $\mathbf{w} \cdot \mathbf{c}_{\text{center}}$. CLP (explicit normalization, black) and CLP-SNN Float (Taylor rule, 64-bit, blue) are robust to all tested $\sigma$ values at both dimensionalities: norms remain near unity and cosine similarity converges to $> 0.95$ even at $\sigma = 0.100$. CLP-SNN INT8 (Loihi~2 integer arithmetic, red) shows increasing norm deviation and slower convergence as $\sigma$ grows, especially at $d = 1280$, due to amplified quantization noise in wide clusters. Nevertheless, the raw dot product $\mathbf{w} \cdot \mathbf{c}_{\text{center}}$ converges to $\approx 1$ for all variants and $\sigma$ values, indicating that the integer rule retains usable prototype alignment even under high intra-class variance. The OpenLORIS features ($d = 1280$, empirically measured effective $\sigma \approx 0.025$--$0.050$) lie within the tested range.}
    \label{fig:learning_rule_sigma_analysis}
\end{figure*}

\begin{figure*}[h!]
    \centering
    \includegraphics[width=0.995\textwidth]{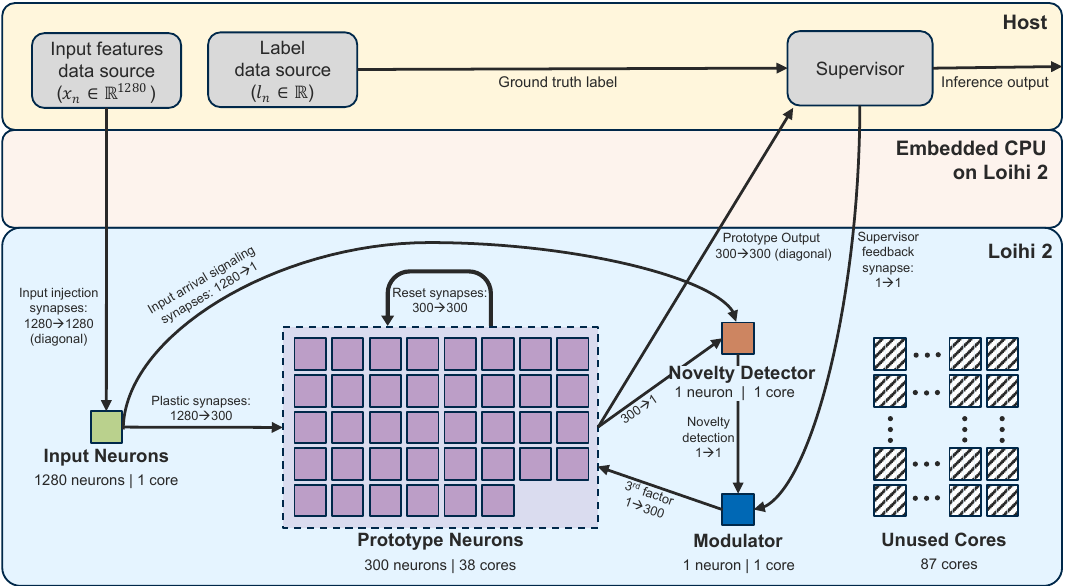}
    \caption{\textbf{Hardware deployment of the CLP-SNN online-learning network on Intel Loihi 2.} Core-map schematic rendered from the Lava partitioner output for the $1280\!\times\!300$ network size used for benchmarking. Cores are color-coded by population (input, prototype, novelty detector and modulator), with arrows, labeled with connection names and shapes, showing spike routing between populations and the third-factor feedback path from the supervisor. The 1280-feature $\times$ 300-prototype CLP-SNN occupies 41 of the 128 neurocores of one Loihi 2 chip; the remaining cores (hatched, right) are idle and available for larger models. Lava's default partitioner places each of the four neuron populations on its own cores, with the number of cores per population set by the tightest on-core memory bank. Input Neurons (1280 neurons, light green) fit on a single core: the upstream input-injection connection is diagonal (1280~$\rightarrow$~1280, non-plastic), so synapse and learning-trace memory remain almost empty. Prototype Neurons (300 neurons, purple) are sharded across 38 cores at $\approx8$ prototypes per core. The dominant constraint is the dense plastic 1280~$\rightarrow$~300 connection: each prototype-hosting core stores the plastic synapses of its prototypes (weight, delay and learning tag) in synapse memory, which shares one memory bank with the table of the 1,581 input axons that every such core receives. The partitioner reduces the number of prototypes per core by an integer factor whenever this bank overflows, which leaves 8 prototypes per core and the prototype cores about half full (49\% of synapse memory). The novelty detector and the modulator neuron each occupy one core. The supervisor and the data sources run on the host computer. The supervisor maps the winning prototype to a class label, compares it with the ground-truth label and sends the result back to the modulator, exchanging spikes with the chip through the embedded CPU (24-bit graded spike messages).}
    \label{fig:loihi_deployment}
\end{figure*}

\begin{figure*}[h!]
    \centering
    \includegraphics[width=0.995\textwidth]{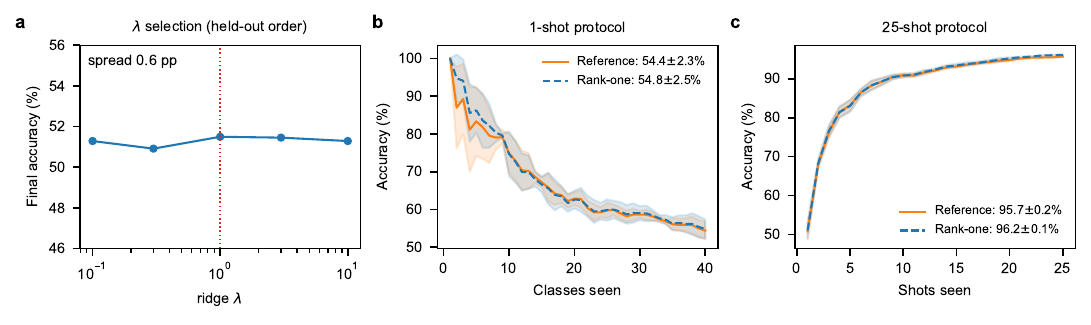}
    \caption{\textbf{The rank-one streaming LDA implementation is accuracy-equivalent to the reference parameterization.} \textbf{a} Final accuracy on a held-out class order (seed-10 1-shot stream, class blocks permuted with an independent random generator) as a function of the fixed ridge $\lambda$; the variation spans 0.6 percentage points across two decades, and $\lambda = 1$ was selected here and used everywhere else. \textbf{b} 1-shot and \textbf{c} 25-shot accuracy trajectories of the reference implementation (shrinkage $\epsilon = 10^{-4}$, inverse recomputed at every checkpoint) and of the rank-one implementation ($\lambda = 1$), mean over three class orders under the evaluation protocols of Table~\ref{table:1}. Final accuracies: 54.4\,$\pm$\,2.3\% versus 54.8\,$\pm$\,2.5\% (1-shot) and 95.7\,$\pm$\,0.2\% versus 96.2\,$\pm$\,0.1\% (25-shot).}
    \label{fig:ea_equivalence}
\end{figure*}

\clearpage
\subsection*{Supplementary Tables}

\begin{table}[h]
\centering
\caption{Cost primitives of the online continual learning step on the
         NVIDIA Jetson Orin Nano (batch size 1, feature dimensionality
         $d = 1280$, 40 classes, unified harness). The \emph{update}
         primitive is the model update on a labeled sample; the
         \emph{query} primitive is scoring one incoming sample. For CLP and
         Replay the update pass already computes the prediction, so no
         separate query enters the step (Table~\ref{table:1}). Latency is
         the mean over all stream samples; the 99th percentile is given
         per primitive and is not additive across primitives. The
         explicit-inversion SLDA rows invert the $1280 \times 1280$
         regularized covariance at every sample; they are a cost anchor
         that quantifies the gain of the rank-one update, not the
         paper's SLDA baseline.
         Substituting the rank-one update removes 21.4~ms of the
         22.9~ms per-sample update on the GPU (94\%).
         In the FP16 rows of the explicit-inversion SLDA the statistics
         are maintained in FP32 and half precision is applied to the
         matrix products only (\hyperref[supp:rankone]{Supplementary Note~3}).}
\label{tab:cost_primitives}
\resizebox{\columnwidth}{!}{%
\begin{tabular}{@{}llc rr rr rr l@{}}
\toprule
 & & & \multicolumn{2}{c}{\textbf{Latency (ms)}} & \multicolumn{2}{c}{\textbf{p99 latency (ms)}} & \multicolumn{2}{c}{\textbf{Dyn.\ energy (mJ)}} & \\
\cmidrule(lr){4-5}\cmidrule(lr){6-7}\cmidrule(lr){8-9}
\textbf{Method} & \textbf{Device} & \textbf{Quant.} & \textbf{upd.} & \textbf{qry.} & \textbf{upd.} & \textbf{qry.} & \textbf{upd.} & \textbf{qry.} & \textbf{Step} \\
\midrule
CLP & GPU & FP32 & 1.39 & --- & 2.23 & --- & 1.15 & --- & update \\  %
CLP & CPU & FP32 & 0.54 & --- & 0.83 & --- & 0.41 & --- & update \\  %
\midrule
NCM & GPU & FP32 & 0.94 & 0.66 & 1.57 & 1.10 & 1.06 & 0.73 & update\,+\,query \\  %
NCM & GPU & FP16 & 0.92 & 0.71 & 1.57 & 1.27 & 1.04 & 0.78 & update\,+\,query \\  %
NCM & CPU & FP32 & 0.45 & 0.28 & 0.50 & 0.46 & 0.41 & 0.20 & update\,+\,query \\  %
\midrule
Replay & GPU & FP32 & 2.41 & --- & 2.98 & --- & 3.11 & --- & update \\  %
Replay & GPU & FP16 & 2.54 & --- & 3.34 & --- & 3.08 & --- & update \\  %
Replay & CPU & FP32 & 5.35 & --- & 10.67 & --- & 16.34 & --- & update \\  %
\midrule
SLDA, explicit inversion & GPU & FP32 & 22.92 & 0.92 & 23.50 & 1.46 & 163.72 & 2.96 & update\,+\,query \\  %
SLDA, explicit inversion & GPU & FP16 & 23.19 & 1.03 & 23.54 & 1.56 & 164.58 & 3.19 & update\,+\,query \\  %
SLDA, explicit inversion & CPU & FP32 & 69.69 & 6.66 & 88.28 & 9.23 & 339.47 & 26.18 & update\,+\,query \\  %
\midrule
SLDA, rank-one & GPU & FP32 & 1.48 & 0.67 & 2.19 & 1.24 & 6.64 & 0.70 & update\,+\,query \\  %
SLDA, rank-one & CPU & FP32 & 5.44 & 0.30 & 9.24 & 0.34 & 20.46 & 0.20 & update\,+\,query \\  %
\bottomrule
\end{tabular}
}
\end{table}

\begin{table}[h]
\centering
\caption{Per-sample operation counts at the benchmark working point
         ($d = 1280$ input dimensions, $C = 40$ classes, batch size~1, 1-shot
         stream). One multiply--accumulate (MAC) is charged
         per scalar product term. \textbf{Top:} inventory for CLP-SNN on Loihi~2
         as the Lava-Loihi compiler instantiates the network ($K = 300$ prototype
         slots, $t_{\mathrm{epoch}} = 20$ time steps per sample, one learning epoch per sample). A
         synaptic operation (one synaptic-memory read for an incoming graded
         spike, with its multiply--accumulate when the weight is non-zero) and one
         evaluation of the plasticity rule at one synapse each count as one
         MAC-equivalent. The input layer emits one graded spike per non-zero
         feature, so the spiking phase costs $\mathrm{nnz}(\mathbf{x})\,K$, where
         $\mathrm{nnz}(\mathbf{x}) = 564 \pm 112$ is the number of non-zero input
         features, measured over the three benchmark streams. Loihi~2 reads every
         stored weight of a dense connection, zeros included, and skips only the
         arithmetic for zero weights~\cite{timcheck2026runtime}, so this term
         covers every stored synapse of each spiking input. The compiler instantiates
         every synapse of the dense plastic $K \times d$ connection, zeros
         included, and configures the learning engine to traverse all of them at
         every epoch, so the learning phase costs $d\,K$ whatever the input
         sparsity; the engine offers pre-synaptic traversal conditions only, so
         the post-synaptic factor of the rule cannot reduce the sweep either. The lateral-inhibition ($K \times K$) and
         third-factor ($K \times 1$) connections are also deployed as plastic
         connections with a rule that changes no weight, because in Lava one neuron cannot have both plastic and non-plastic synapses. The engine sweeps them as well, so they are listed separately as deployment overhead. Dividing the measured
         dynamic energy per step (0.013~mJ, rounded to 0.01~mJ in
         Table~\ref{table:1}) by the deployed count gives about 20~pJ per
         MAC-equivalent. \textbf{Bottom:} per-step MACs of
         the conventional learners on the Orin under the composition rule of
         Table~\ref{table:1}, with their measured step latencies. CLP computes
         only against its allocated prototypes, so its count is evaluated at the
         stream-mean allocation of the benchmark stream ($M = 65.8$) and grows
         with the number of allocated prototypes, as does the chip's count with
         the number of instantiated slots; the counts of NCM, Replay and SLDA are
         fixed. Replay uses $B = 51$ samples per update (the incoming sample plus
         50 replayed). The explicit-inversion SLDA row is the cost anchor of
         Supplementary Table~\ref{tab:cost_primitives}. Across the four reported
         learners a $97\times$ range of operation counts produces a $1.7\times$
         range of GPU step latency and an $11\times$ range on the CPU.}
\label{tab:op_counts}
\resizebox{\columnwidth}{!}{%
\begin{tabular}{@{}llrrr@{}}
\toprule
\multicolumn{5}{@{}l}{\textbf{CLP-SNN on Loihi 2 (per sample)}} \\
\textbf{Term} & \textbf{Symbolic} & \textbf{Count} & & \\
\midrule
Feedforward spike fan-out (spiking phase) & $\mathrm{nnz}(\mathbf{x})\,K$ & 169{,}350 & & \\
Plasticity sweep, prototype synapses (learning phase) & $d\,K$ & 384{,}000 & & \\
Control plane (WTA reset, novelty, allocation, feedback) & $\approx 2\,\mathrm{nnz}(\mathbf{x}) + 3K$ & 2{,}032 & & \\
\textbf{Step, algorithm} & $\mathrm{nnz}(\mathbf{x})\,K + d\,K + \text{control}$ & \textbf{555{,}382} & & \\
Plasticity sweep, lateral-inhibition and third-factor connections & $K^{2} + K$ & 90{,}300 & & \\
\textbf{Step, as deployed} & & \textbf{645{,}682} & & \\
\midrule
\multicolumn{5}{@{}l}{\textbf{Conventional learners on the Orin (per step)}} \\
\textbf{Method} & \textbf{Symbolic} & \textbf{MAC} & \textbf{GPU (ms)} & \textbf{CPU (ms)} \\
\midrule
NCM & $Cd + 2d$ & $5.4 \times 10^{4}$ & 1.60 & 0.73 \\
CLP ($M = 65.8$) & $Md + d$ & $8.6 \times 10^{4}$ & 1.39 & 0.54 \\  %
SLDA, rank-one & $2d^{2} + 3dC$ & $3.4 \times 10^{6}$ & 2.15 & 5.74 \\
Replay ($B = 51$) & $2BdC$ & $5.2 \times 10^{6}$ & 2.41 & 5.35 \\
SLDA, explicit inversion (anchor) & $d^{3} + (C+1)d^{2}$ & $2.2 \times 10^{9}$ & 23.8 & 76.3 \\
\bottomrule
\end{tabular}
}
\end{table}

\begin{table}[h]
\centering
\caption{CLP-SNN hyperparameters used in this work. The hardware values of $v_{\mathrm{th}}$ and $\beta$ satisfy the constraint $\theta = y_{\min} = v_{\mathrm{th}} - t_{\mathrm{wait}}\,\beta$ (Eq.~\ref{eq:novelty_threshold} of main text); the float and INT8 simulators parameterize this directly via the effective threshold $\theta$, while the on-chip deployment selects integer $v_{\mathrm{th}}$ and $\beta$ consistent with this constraint.}
\label{tab:hyperparameters}
\begin{tabular}{ll>{\raggedright\arraybackslash}p{0.54\textwidth}}
\toprule
\textbf{Parameter} & \textbf{Value} & \textbf{Description} \\
\midrule
$\theta\;(\equiv y_{\min})$  & $0.71$ / $0.51$     & Effective cosine-similarity threshold (1-shot / 25-shot) \\
$t_{\mathrm{wait}}$          & $11$                & \# of SNN time steps for novelty-detector countdown window \\
$t_{\mathrm{epoch}}$         & $20$                & \# of SNN time steps per learning epoch (one epoch per input sample)\\
$K$                          & $300$               & Pre-allocated prototype slots (benchmark network; the accuracy runs size the pool per protocol) \\
$d$                          & $1280$              & Input feature dimensionality \\
Number of classes            & $40$                & OpenLORIS category count \\
$g^{(k)}(0)$                 & $1$                 & Initial goodness score \\
$\Delta g$                   & $0.5$               & Goodness step per update, in simulation (0.2 in Supplementary Fig.~\ref{fig:learning_rule_sigma_analysis}) \\
$\alpha^{(k)}(0)$            & $1$                 & Initial learning rate ($=1/g^{(k)}(0)$) \\
INT8 weight range            & $[-128,\,127]$      & 8-bit signed, saturating clamp \\
INT8 input scale             & $S = 128$           & $x_{\mathrm{int}} = \mathrm{round}(x\cdot S)$ \\
Weight-update rounding       & Stochastic          & Shared scalar draw per update; 15-bit accumulator $\to$ 8-bit \\
Independent runs / experiment & $3$                & Different random seeds for class order and choice of video clips \\
\bottomrule
\end{tabular}
\end{table}

\clearpage
\subsection*{Supplementary Note 1: Sources of the accuracy gap between CLP and CLP-SNN}
\phantomsection\label{supp:gap}

The CLP-SNN accuracies in Table~\ref{table:1} and Fig.~\ref{fig:exp_results} are measured on Loihi~2, where the network imprints each new prototype and does not adapt it afterwards. To locate the sources of the accuracy gap to CLP, we emulated this deployed network in integer arithmetic. Inputs are clipped to non-negative values, normalized and rounded to 7 bits, and a new prototype stores the rounded input. Prototypes spike earlier the more similar they are to the input; one time step spans a cosine similarity of 0.024 (1-shot) or 0.043 (25-shot), and prototypes that spike in the same step vote on the label. During training a prototype wins only if its similarity reaches the deployed threshold, 0.71 (1-shot) or 0.51 (25-shot). No parameter is fitted to the chip's results. Evaluated as on the chip, the emulator reproduces the measured final accuracy to within 0.2 percentage points, and its accuracy curves deviate from the measured ones by 0.4 to 0.5 points (root mean square).

Supplementary Table~\ref{tab:gap} adds the features of the deployment to CLP one at a time. Imprinting prototypes without adapting them, with 8-bit arithmetic, matches the accuracy of CLP at CLP's threshold, but only by allocating about twice as many prototypes, and a larger network costs latency and energy on Loihi~2. The deployment therefore uses a lower similarity threshold, which keeps the network small and accounts for most of the accuracy gap. Resolving the winner in discrete spike-time steps costs a further 0.5 points in the 1-shot setting. Adapting prototypes, as the full CLP-SNN rule does, avoids the prototype overhead of imprint-only operation (\hyperref[supp:adaptive]{Supplementary Note~2}).

\begin{table}[h]
\centering
\caption{Final accuracy (\%) and number of prototypes (in parentheses) as the features of the Loihi~2 deployment are added to CLP one at a time, computed with the deployment emulator. Mean over three class orders.}
\label{tab:gap}
\begin{tabular}{lrr}
\toprule
\textbf{Configuration} & \textbf{1-shot} & \textbf{25-shot} \\
\midrule
CLP (Table~\ref{table:1})                                   & 57.0 (127) & 93.0 (1{,}769) \\
Imprint-only, 8-bit, CLP's threshold (0.75)                 & 56.9 (242) & 92.8 (3{,}673) \\
\quad + input clipping                                      & 57.2 (210) & 92.5 (3{,}083) \\
\quad + deployed threshold (0.71 / 0.51)                    & 55.9 (167) & 90.4 (1{,}217) \\
\quad + spike-time winner selection                         & 55.4 (168) & 90.5 (1{,}302) \\
\bottomrule
\end{tabular}
\end{table}

\clearpage
\subsection*{Supplementary Note 2: Adaptive CLP-SNN in simulation}
\phantomsection\label{supp:adaptive}

The full CLP-SNN rule also updates the winning prototype after correct and incorrect predictions, with a learning rate set by its goodness score. It currently runs only in simulation. Supplementary Table~\ref{tab:adaptive} compares it with CLP and with imprint-only operation across similarity thresholds. In floating point, adaptive CLP-SNN allocates fewer prototypes than CLP and about half as many as imprint-only operation, and stays within 1.8 points of CLP's accuracy. With 8-bit weights the adaptive rule loses a further 1.5 to 6.2 points, most in the 1-shot setting at low thresholds. The learning rate $1/g$ decreases as a prototype accumulates correct predictions, and after a few updates the change of most weights falls below one step of the 8-bit weight ($1/128$), which limits how precisely an 8-bit prototype can average its samples. At low thresholds each prototype absorbs more samples and reaches this regime sooner. Imprint-only operation is unaffected, because it writes each weight once.

\begin{table}[h]
\centering
\caption{Final accuracy (\%) and number of prototypes (in parentheses) of CLP and of CLP-SNN in simulation across similarity thresholds. Adaptive: the full CLP-SNN rule; imprint-only: prototypes are not updated after allocation. Mean over three class orders.}
\label{tab:adaptive}
\begin{tabular}{llrrrr}
\toprule
 & & \multicolumn{4}{c}{\textbf{Similarity threshold}} \\
\cmidrule(lr){3-6}
\textbf{Setting} & \textbf{Method} & \textbf{0.70} & \textbf{0.75} & \textbf{0.80} & \textbf{0.85} \\
\midrule
1-shot   & CLP                    & 56.6 (99)  & 57.0 (127) & 56.8 (183) & 57.2 (281) \\
         & Adaptive, FP32         & 55.0 (80)  & 56.1 (104) & 56.6 (145) & 56.4 (217) \\
         & Adaptive, 8-bit        & 48.9 (87)  & 51.4 (107) & 52.4 (159) & 54.5 (243) \\
         & Imprint-only, 8-bit    & 56.4 (191) & 56.9 (242) & 56.6 (328) & 57.2 (455) \\
\midrule
25-shot  & CLP                    & 92.3 (1{,}127) & 93.0 (1{,}769) & 93.1 (2{,}847) & 93.2 (4{,}831) \\
         & Adaptive, FP32         & 90.5 (1{,}039) & 91.6 (1{,}554) & 92.5 (2{,}405) & 93.2 (3{,}983) \\
         & Adaptive, 8-bit        & 88.0 (1{,}263) & 90.1 (1{,}839) & 90.3 (2{,}771) & 91.7 (4{,}614) \\
         & Imprint-only, 8-bit    & 92.4 (2{,}670) & 92.8 (3{,}673) & 92.9 (5{,}265) & 93.2 (7{,}902) \\
\bottomrule
\end{tabular}
\end{table}
\clearpage
\subsection*{Supplementary Note 3: Exact rank-one maintenance of the streaming LDA precision}
\phantomsection\label{supp:rankone}

\paragraph{Setup and invariants.}
Symbols follow the Methods subsection \hyperref[sec:methods_rankone_slda]{``Exact rank-one streaming LDA baseline''}. The state is the precision $\boldsymbol{\Lambda}_n = \mathbf{A}_n^{-1}$ of the regularized scatter $\mathbf{A}_n = \mathbf{S}_n + \lambda\mathbf{I}$, the class means $\mathbf{M}_n$ with counts $n_\ell$, the discriminant weights $\boldsymbol{\Omega}_n = \boldsymbol{\Lambda}_n\mathbf{M}_n$, the biases $b_\ell = \tfrac12\boldsymbol{\mu}_\ell^{\top}\boldsymbol{\omega}_\ell$, and the global counter $n$. Initialization: $\mathbf{A}_0 = \lambda\mathbf{I}$, $\boldsymbol{\Lambda}_0 = \mathbf{I}/\lambda$, $\boldsymbol{\Omega}_0 = \mathbf{0}$, $b = \mathbf{0}$, $n = 0$. The covariance update of the reference implementation~\cite{hayes2020deepSLDA}, Eq.~(\ref{eq:slda_cov}), becomes Eq.~(\ref{eq:slda_rank1}) in scatter form; the first sample has $c_0 = 0$ and leaves the scatter unchanged. Two invariants hold by induction: $\mathbf{A}_n \succ 0$, because a positive-semidefinite term is added at every step and nothing is ever subtracted, and $\boldsymbol{\Lambda}_n$ is symmetric, because each update subtracts a symmetric rank-one term, $\gamma_n\mathbf{h}_n\mathbf{h}_n^{\top}$ (defined below). The implementation therefore needs no explicit symmetrization step.

\paragraph{Exactness.}
Let $\mathbf{A} \succ 0$ with $\boldsymbol{\Lambda} = \mathbf{A}^{-1}$, and let $\mathbf{A}' = \mathbf{A} + c\,\mathbf{z}\mathbf{z}^{\top}$ with $c \geq 0$. Define $\mathbf{h} = \boldsymbol{\Lambda}\mathbf{z}$, $q = \mathbf{z}^{\top}\mathbf{h}$ and $\gamma = c/(1 + cq)$. Then $\boldsymbol{\Lambda}' = \boldsymbol{\Lambda} - \gamma\,\mathbf{h}\mathbf{h}^{\top}$ satisfies
\begin{align}
\mathbf{A}'\boldsymbol{\Lambda}'
&= (\mathbf{A} + c\,\mathbf{z}\mathbf{z}^{\top})(\boldsymbol{\Lambda} - \gamma\,\mathbf{h}\mathbf{h}^{\top}) \nonumber\\
&= \underbrace{\mathbf{A}\boldsymbol{\Lambda}}_{=\,\mathbf{I}}
 \;-\; \gamma\underbrace{(\mathbf{A}\mathbf{h})}_{=\,\mathbf{z}}\mathbf{h}^{\top}
 \;+\; c\,\mathbf{z}\underbrace{(\mathbf{z}^{\top}\boldsymbol{\Lambda})}_{=\,\mathbf{h}^{\top}}
 \;-\; c\gamma\,\mathbf{z}\underbrace{(\mathbf{z}^{\top}\mathbf{h})}_{=\,q}\mathbf{h}^{\top} \nonumber\\
&= \mathbf{I} + \bigl[c - \gamma(1 + cq)\bigr]\mathbf{z}\mathbf{h}^{\top} = \mathbf{I},
\end{align}
where $\mathbf{A}\mathbf{h} = \mathbf{A}\boldsymbol{\Lambda}\mathbf{z} = \mathbf{z}$ and $\mathbf{z}^{\top}\boldsymbol{\Lambda} = \mathbf{h}^{\top}$ by the symmetry of $\boldsymbol{\Lambda}$. The bracket vanishes identically because $\gamma = c/(1 + cq)$ is the unique coefficient that cancels the cross term. By uniqueness of the inverse, $\boldsymbol{\Lambda}' = (\mathbf{A}')^{-1}$ exactly~\cite{sherman1950adjustment,bartlett1951inverse}; no approximation is involved. Since $q = \mathbf{z}^{\top}\boldsymbol{\Lambda}\mathbf{z} \geq 0$ and $c \in [0, 1)$, the denominator satisfies $1 + cq \geq 1$, so no small-divisor amplification is possible and $\gamma < 1$ bounds the subtracted patch by the scale of $\mathbf{h}\mathbf{h}^{\top}$. The implementation asserts $1 + c_n q_n \geq 1$ at every step in its test mode.

\paragraph{The gain identity.}
From $\boldsymbol{\Lambda}' = \boldsymbol{\Lambda} - \gamma\,\mathbf{h}\mathbf{h}^{\top}$ and $\mathbf{h} = \boldsymbol{\Lambda}\mathbf{z}$,
\begin{equation}
\boldsymbol{\Lambda}'\mathbf{z} = \mathbf{h} - \gamma\,\mathbf{h}\,(\mathbf{h}^{\top}\mathbf{z}) = \mathbf{h}\,(1 - \gamma q) = \mathbf{h}\cdot\frac{1 + cq - cq}{1 + cq} = \frac{\mathbf{h}}{1 + cq},
\end{equation}
which has the form of the gain vector of recursive least squares~\cite{plackett1950theorems}. It supplies the $O(d)$ column correction of Eq.~(\ref{eq:slda_meanshift}) without a second matrix--vector product. The bias entries follow from the two weight patches: the precision patch changes column $\ell'$ by $-\gamma\,(\mathbf{M}^{\top}\mathbf{h})_{\ell'}\,\mathbf{h}$, hence $\Delta b_{\ell'} = -\tfrac12\gamma\,(\mathbf{M}^{\top}\mathbf{h})_{\ell'}^{2}$ for all $\ell'$ at a cost of $O(C)$; the mean and column update of class $\ell$ changes only $b_\ell$, which is recomputed exactly as $\tfrac12\boldsymbol{\mu}_\ell^{\top}\boldsymbol{\omega}_\ell$ at a cost of $O(d)$.

\paragraph{Pseudocode.}
Algorithm~\ref{alg:rankone_slda} mirrors the released implementation line for line; the order matters: $\mathbf{z}$ must be computed against the old mean, and the precision and weights must be patched before the mean moves.

\begin{algorithm}[h]
\caption{Exact rank-one streaming LDA: one update and one prediction}
\label{alg:rankone_slda}
\begin{algorithmic}[1]
\Require state $\boldsymbol{\Lambda}$ ($d\times d$), $\mathbf{M}$ ($d\times C$), counts $n_\ell$, $\boldsymbol{\Omega}$ ($d\times C$), $b$ ($C$), counter $n$
\Procedure{Fit}{$\mathbf{x}, \ell$}
  \State $\mathbf{z} \gets \mathbf{x} - \mathbf{M}_{:,\ell}$;\quad $c \gets n/(n+1)$
  \State $\mathbf{h} \gets \boldsymbol{\Lambda}\mathbf{z}$;\quad $q \gets \mathbf{z}^{\top}\mathbf{h}$;\quad $\gamma \gets c/(1 + cq)$
  \If{$c > 0$}
    \State $\boldsymbol{\Lambda} \gets \boldsymbol{\Lambda} - \gamma\,\mathbf{h}\mathbf{h}^{\top}$ \Comment{precision patch, Eq.~(\ref{eq:slda_sm})}
    \State $\mathbf{s} \gets \mathbf{M}^{\top}\mathbf{h}$;\quad $\boldsymbol{\Omega} \gets \boldsymbol{\Omega} - \gamma\,\mathbf{h}\mathbf{s}^{\top}$;\quad $b \gets b - \tfrac12\gamma\,\mathbf{s}\odot\mathbf{s}$ \Comment{weights and biases, Eq.~(\ref{eq:slda_wpatch})}
  \EndIf
  \State $\mathbf{M}_{:,\ell} \gets \mathbf{M}_{:,\ell} + \mathbf{z}/(n_\ell + 1)$;\quad $n_\ell \gets n_\ell + 1$ \Comment{class mean}
  \State $\boldsymbol{\Omega}_{:,\ell} \gets \boldsymbol{\Omega}_{:,\ell} + \mathbf{h}/\bigl((1 + cq)\,n_\ell\bigr)$ \Comment{gain identity, Eq.~(\ref{eq:slda_meanshift}); $n_\ell$ is already incremented}
  \State $b_\ell \gets \tfrac12\,\mathbf{M}_{:,\ell}^{\top}\boldsymbol{\Omega}_{:,\ell}$;\quad $n \gets n + 1$
\EndProcedure
\Procedure{Predict}{$\mathbf{x}$}
  \State \Return $\arg\max_{\ell:\,n_\ell > 0}\left(\mathbf{x}^{\top}\boldsymbol{\Omega} - b\right)_\ell$ \Comment{one $d\times C$ product; no inversion}
\EndProcedure
\end{algorithmic}
\end{algorithm}

\paragraph{Regularization equivalence.}
The reference implementation inverts $(1-\epsilon)\boldsymbol{\Sigma}_n + \epsilon\mathbf{I}$ with $\epsilon = 10^{-4}$. Since $\boldsymbol{\Sigma}_n = \mathbf{S}_n/n$, this equals $\tfrac{1-\epsilon}{n}\bigl[\mathbf{S}_n + \tfrac{\epsilon n}{1-\epsilon}\mathbf{I}\bigr]$: a ridge on the scatter that grows linearly with $n$, which is not a rank-one change and cannot be tracked by Eq.~(\ref{eq:slda_sm}). The fixed ridge $\mathbf{S}_n + \lambda\mathbf{I}$ corresponds to a shrinkage intensity $\lambda/n$ that fades as evidence accumulates. Because the decision is an argmax over scores linear in $\boldsymbol{\Lambda}$, a positive rescaling of $\boldsymbol{\Lambda}$ cannot change the winner, so the two parameterizations differ only in their shrinkage schedule.

Supplementary Fig.~\ref{fig:ea_equivalence} reports the check: $\lambda$ was selected once on a held-out class order (the seed-10 1-shot stream with its class blocks permuted by an independent random generator), where final accuracy lies between 50.9\% and 51.5\% across $\lambda \in \{0.1, 0.3, 1, 3, 10\}$; with $\lambda = 1$, the rank-one implementation reaches 54.8\,$\pm$\,2.5\% (1-shot) and 96.2\,$\pm$\,0.1\% (25-shot) against 54.4\,$\pm$\,2.3\% and 95.7\,$\pm$\,0.2\% for the reference parameterization, over the same three class orders and evaluation protocols as Table~\ref{table:1}. The substitution therefore does not weaken the baseline.

\paragraph{Numerical checks.}
Three checks support exactness and numerical stability in practice. In FP64 the maintained $\boldsymbol{\Lambda}$ matches a from-scratch inverse of $\mathbf{S} + \lambda\mathbf{I}$ to a relative deviation below $10^{-11}$ at every checkpoint. In FP32 over the 2,400-sample 1-shot stream, the relative deviation of $\boldsymbol{\Lambda}$ from that reference stays below $2.7\times10^{-6}$ and that of the class scores below $2.4\times10^{-6}$, with 100\% agreement on the predicted labels of the full test set. End to end, the 25-shot comparison above, over a stream of about 60,000 samples, shows no accumulation-driven degradation. This is what the stability argument above predicts: with $1 + c_n q_n \geq 1$ at every step, the recursion has no error-amplifying divisor, and the patch magnitude is bounded by $\gamma_n < 1$. The state is kept in FP32 throughout.

\paragraph{Cost primitives and the explicit-inversion anchor.}
Supplementary Table~\ref{tab:cost_primitives} reports both cost primitives for every method under the unified harness, including the half-precision variants and an explicit-inversion streaming LDA implementation that recomputes the precision with an LU factorization at every sample. The explicit-inversion implementation performs about $2\times10^{9}$ multiply--accumulate operations per sample against $3.4\times10^{6}$ for the rank-one update, and holds both the covariance and the precision; it is included as the cost anchor that quantifies what the rank-one update saves, not as a baseline in its own right. Half precision does not help it, because the inversion itself runs in FP32. More generally, half-precision execution of streaming LDA is only viable with an FP32 accumulator state: the relative increments of the covariance recursion scale as $1/n$ and fall below FP16 resolution as the stream grows, so an FP16 covariance state degrades progressively and its agreement with the FP32 model falls to 1.6\% by the end of the 1-shot stream, whereas FP16 matrix products with FP32 state keep agreement above 99.9\% at every checkpoint. The rank-one implementation keeps its state in FP32 for the same reason, and no half-precision variant of it is reported.

\end{document}